\documentclass{article}
\usepackage{arxiv}
\usepackage{nicefrac}       
\usepackage{microtype}      
\usepackage{booktabs}       
\usepackage{doi}
\usepackage{adjustbox}
\usepackage{float}
\usepackage[T1]{fontenc}
\usepackage{graphicx}
\usepackage{hhline}
\usepackage[utf8]{inputenc}
\usepackage{subcaption}
\usepackage{multicol}
\usepackage{multirow}
\usepackage[normalem]{ulem}
\usepackage{hyperref}       
\usepackage{url}            
\usepackage{amsfonts}       
\usepackage{graphicx}
\usepackage{authblk}
\usepackage{flafter}
\usepackage[square,sort,comma,numbers]{natbib}

\title{Machine Learning-Assisted Recurrence Prediction for Early-Stage Non-Small-Cell Lung Cancer Patients}

\author[1]{Adrianna Janik} 
\author[2]{Maria Torrente}
\author[1]{Luca Costabello}
\author[2]{Virginia Calvo}
\author[3,4]{Brian Walsh}
\author[5]{Carlos Camps }
\author[3,4]{Sameh K. Mohamed}
\author[6]{Ana L. Ortega}
\author[3,4,9,10]{Vít Nováček}
\author[7]{Bartomeu Massutí}
\author[8]{Pasquale Minervini}
\author[11]{M.Rosario Garcia Campelo}
\author[12]{Edel del Barco}
\author[13]{Joaquim Bosch-Barrera}
\author[14]{Ernestina Menasalvas}
\author[3,4]{Mohan Timilsina}
\author[2]{Mariano Provencio}

\affil[1]{Accenture Labs, Dublin, Ireland}
\affil[2]{Medical Oncology Department, Hospital Universitario Puerta de Hierro Majadahonda, Madrid, Spain}
\affil[3]{Data Science Institute, University of Galway, Galway, Ireland}
\affil[4]{Insight Centre for Data Analytics, University of Galway, Galway, Ireland}
\affil[5]{Hospital General de Valencia, Valencia, Spain}
\affil[6]{Hospital Universitario de Jaén, Jaen, Spain}
\affil[7]{Hospital General Universitario de Alicante, Alicante, Spain}
\affil[8]{University College London, London, United Kingdom}
\affil[9]{Faculty of Informatics, Masaryk University, Brno, Czech Republic}
\affil[10]{Masaryk Memorial Cancer Institute, Brno, Czech Republic}
\affil[11]{Complejo Hospitalario Universitario A Coruña , A Coruna, Spain}
\affil[12]{Hospital Universitario de Salamanca, Salamanca, Spain}
\affil[13]{Institut Català d’Oncologia, Hospital Universitari Dr. Josep Trueta, Girona, Spain}
\affil[14]{Polytechnic University of Madrid, Madrid, Spain}

\renewcommand{\shorttitle}{ML-Assisted Recurrence Prediction for Early-Stage NSCLC Patients}
\date{}

\hypersetup{
pdftitle={Machine Learning-Assisted Recurrence Prediction for Early-Stage Non-Small-Cell Lung Cancer Patients},
pdfsubject={},
pdfauthor={},
pdfkeywords={},
}

\begin{document}
\maketitle

\begin{abstract}
	\textbf{Background:} Stratifying cancer patients according to risk of relapse can personalize their care. In this work, we provide an answer to the following research question: How to utilize machine learning to estimate probability of relapse in early-stage non-small-cell lung cancer patients?

\textbf{Methods:} For predicting relapse in 1,387 early-stage (I-II), non-small-cell lung cancer (NSCLC) patients from the Spanish Lung Cancer Group data (65.7 average age, 24.8\% females, 75.2\% males) we train tabular and graph machine learning models. We generate automatic explanations for the predictions of such models. For models trained on tabular data, we adopt SHAP local explanations to gauge how each patient feature contributes to the predicted outcome. We explain graph machine learning predictions with an example-based method that highlights influential past patients.

\textbf{Results:} Machine learning models trained on tabular data exhibit a 76\% accuracy for the Random Forest model at predicting relapse evaluated with a 10-fold cross-validation (model was trained 10 times with different independent sets of patients in test, train and validation sets, the reported metrics are averaged over these 10 test sets). Graph machine learning reaches 68\% accuracy over a 200-patient, held-out test set, calibrated on a held-out set of 100 patients.

\textbf{Conclusions:} Our results show that machine learning models trained on tabular and graph data can enable objective, personalised and reproducible prediction of relapse and therefore, disease outcome in patients with early-stage NSCLC. With further prospective and multisite validation, and additional radiological and molecular data, this prognostic model could potentially serve as a predictive decision support tool for deciding the use of adjuvant treatments in early-stage lung cancer. 
\end{abstract}

\section{Introduction}
Lung cancer is the world’s leading cause of cancer-related death with an estimated 1.8 million deaths equating to 18$\%$ of all cancer deaths in 2020 \cite{sung_global_2021}. It is the leading cause of cancer-related deaths for men and second for women after breast cancer \cite{sung_global_2021}. Mortality and incidence occur roughly twice as much in men than in women \cite{sung_global_2021}. Five years relative survival after being diagnosed with lung and bronchus cancer in US population was 22$\%$ as reported in Cancer Statistics 2022 by ACS for years 2011–2017 \cite{noauthor_cancer_nodate}. 

The resection of early-stage NSCLC offers patients the best hope of a cure, however, relapse rates post-resection remain high and even for patients with disease at the same stage incidences of relapse after curative surgery vary significantly. 30$\%$ to 55$\%$ of patients with NSCLC develop relapse and eventually die of their disease despite curative resection.  Therefore, accurately predicting the individual patient cases in which the disease is likely to recur after surgery can facilitate early, personalised detection and treatment \cite{uramoto_recurrence_2014}.

This work targets machine-aided prediction of lung cancer relapse within the scope of the European project CLARIFY\footnote{ https://www.clarify2020.eu/}, an effort focused on monitoring health status and quality of life after cancer treatment. It is a follow-up of \cite{sameh_k_mohamed_et_al_predicting_2021} in which we explored baseline models for NSCLC patients elapse prediction. Since the capacity of baseline models for making complex inferences based on individual patients’ data is rather limited, in this work we model clinical data as a knowledge graph \cite{hogan_knowledge_2022}. Compared to modelling data as tables, a knowledge graph  favours extensions with heterogenous data sources \cite{wilcke_knowledge_2017}. Graph datasets are well suited to be used by various reasoners: logical, machine learning-based, or hybrid and, allow inference over long-range dependencies between concepts (e.g., patients, diseases, biomarkers, genes, drugs). Graphs also ease  integrating core clinical data with population-level genetics, drugs, diseases \cite{pinero_disgenet_2019, fabregat_reactome_2018, mendez_chembl_2019}, and information extracted from relevant scientific literature. Graph machine learning adopted in this work leverages dependencies and patterns between patients and other concepts in the graph and delivers more powerful predictions over traditional approaches. We complement  existing approaches in survival analysis applied to relapse prediction (e.g. machine learning used  to estimate Time-To-Relapse (TTR) for malignant pleural mesothelioma \cite{zauderer_use_2021}, or studies that analyze prognostic features for NSCLC relapse tumor tissue image analysis \cite{corredor_spatial_2019, huynh_associations_2017,wang_prediction_2017}). Unlike the previous work \cite{lee_deepbts_2020} that used time-binned deep neural networks, we also propose a novel method to explain each machine learning prediction by retrieving influential patients from the training set.

\begin{figure}[htbp!]
    \centering
\includegraphics[width=16.51cm,height=9.53cm]{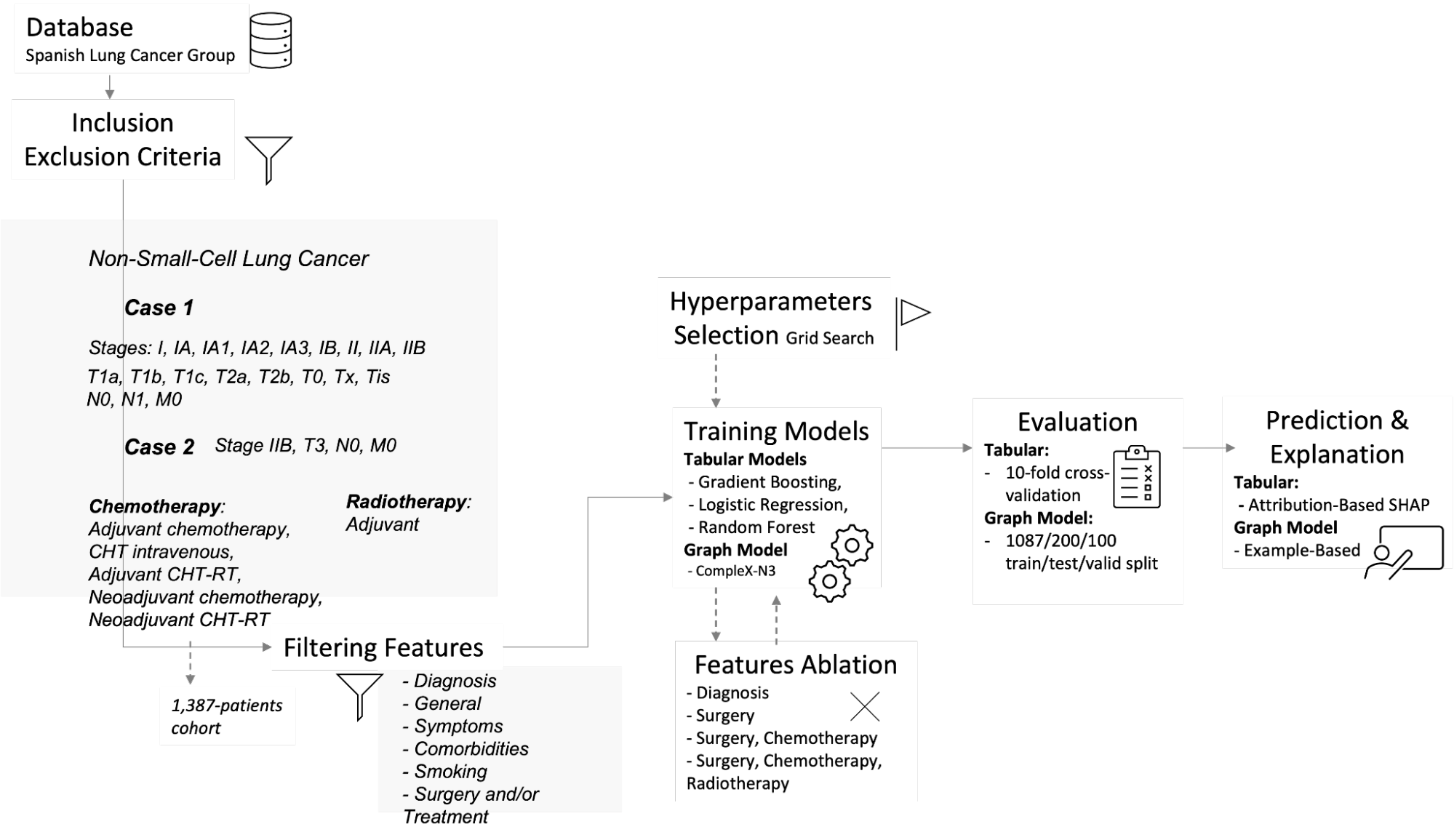}
    \caption{Diagram representing the prediction pipeline, from the database, through criteria and features, models training, features ablation, evaluation until predicting and explaining.}
    \label{fig:my_label}
\end{figure}

\section{Materials and Methods}
\textbf{Patients.} We carried out a retrospective study looking at 1,387 I-II early-stage NSCLC patients selected according to the criteria in \textit{Table 1} from the Spanish Lung Cancer Group’s Thoracic Tumor Registry (TTR) containing at the time of writing 12,981 NSCLC patients. The 1,387 patients come from two cases filtered according to treatment type received. \textit{Case 1 }(n$=$1,711) tumor stage I and II with stage T: T1a/T1b/T1c/T2a/T2b/T0/Tx/Tis; stage N: N0/N1; and M: M0. \textit{Case 2 }(n$=$375)\textit{ }tumor stage of IIB; TNM stage: T3, N0, M0. Apart from tumor characteristics, we filtered patients who underwent one of the following chemotherapy types: Adjuvant chemotherapy/intravenous CHT/Adjuvant CHT-RT/Neoadjuvant chemotherapy/Neoadjuvant CHT-RT/none; and/or Adjuvant radiotherapy.

\textbf{Study design } We treat predicting risk of relapse as a binary classification task and further cast it as a tabular model classification and graph-based link prediction tasks as depicted in Figure 1. These two approaches differ in the data representation aspect (see Introduction). We train two sets of models on the same input data, in a table-like andgraph forms. We evaluate models using binary classification metrics, adopting a 10-fold cross-validation protocol for tabular models and train/valid/test splits for graph model.

\textbf{Labels.} We label patients as positives if they either i) have a progression record with status "Progression"/"Relapse" or ii) their follow-up records include the status "Alive with disease"/"Dead" (with cause of death "Lung cancer") similarily to \cite{sameh_k_mohamed_et_al_predicting_2021}. The rest are considered negatives. There were 37.3$\%$ (n$=$517) relapsed patients in any time of their life after initial surgery and/or treatment and 62.7$\%$ (n$=$870) patients who did not. 

\textbf{Patients' features.} We identify feature as both patient and treatment characteristics. For full list of features with descriptions see supplement-section:D. We trained models on the group of features: tumor characteristics (stage, TNM, histology,grade, and subtype details), general (age, race, gender, previous cancer type, family cancer history, ECOG, synchronous tumors, biomarkers: ALK IHQ, PD-L1, EGFR negative), comorbidities, smoking information, symptoms, radiotherapy (type, area, dose, fractioning, duration), chemotherapy (type, start time, regimen), surgery (procedure, time, type, resection grade, TN stages).

\textbf{Data Preprocessing.} The original dataset includes several dates which we turn into months-elapsed-since-the-diagnosis-date. If the diagnosis date is not present, we choose one month before the first recorded treatment as the estimated date. Each time-stamped feature is derived by subtracting its date from the above timestamp. This applies to chemotherapy start time, surgery time. Radiotherapy duration is in days. Patient age is derived based on the heuristic estimating age-at-diagnosis, when the diagnosis date is missing, it estimates it based on the first available date (e.g., oncological consultation). Additional condition cuts dates more recent than 1990. For the graph machine learning model, we group numerical features into range bins (the graph machine learning model adopted in this work does not support continuous numerical features). Table 2 presents patientscharacteristics.

\textbf{Machine Learning Models Training} Both models tabular and graph-based share a similar pipeline depicted in Figure 1. We found the best hyperparameters using grid search over predefined values and used the best combinations for training models on different subsets of features in ablation studies (results reported in Table 3). The first model was trained only on diagnosis and general features and then two models were trained with addition of the following groups of features: 

\begin{itemize}
	\item Surgery: type, time-since-diagnosis (months), procedure, resection grade, T,Nstages.

	\item Surgery (as above) $\&$ Chemotherapy: type, start-time-since-diagnosis (months), regimen $\&$Radiotherapy: type, duration (in days), dose, fractioning, area.

\end{itemize}

\textbf{\textit{Relapse Prediction with Tabular Machine Learning}}

To predict relapse, we first represent patients features in a table-like form and adopt machine learning classifiers:logistic regression, random forest, and gradient boosting\footnote{ We exclude the multi-layer perceptron and support vector classifier (present in[4]) due to their lack of interpretability, lower predictive power, and higher training time.}. We trained these three models on the patient data. Each model was trained for three different sets of features as described above. Models predict whether a patient is going to relapse or not. Preprocessing applied to the data includes imputation of missing values with constant value\footnote{ -1 for non-time related features and 5000 months for time-related features.} for numerical features and a special code for categorical features. The latter were then one-hot-encoded. Because of the imbalanced nature of the data, we randomly over-sampled. 

The hyperparameters of the models are chosen using a grid search procedure over the predefined set of parameters grid for each type of model. Models were trained using stratified 10-fold cross-validation where each set contains approximately the same percentage of each target class. We choose hyperparameters corresponding to the best performing configuration for each of the examined models. The best hyperparameters are reported in the supplementary material.

\textbf{\textit{Relapse Prediction with Graph Machine Learning}}

\textbf{Dataset.} We model the\textbf{ }lung cancer clinical data as a knowledge graph \cite{hogan_knowledge_2022} that includes 34,351 statements that connect nodes in the graph (42 links types, 1524 nodes). The graph covers 1,387 early-stage NSCLC patients and their clinical data, see Figure 2. 

Test and validation patients are selected randomly, enforcing a balanced distribution of classes in the test and validation sets. The number of patients is  pre-set to 200 in the test set ($\sim$14.8$\%$) and 100 in the validation set ($\sim$7,4$\%$). Respectively, there are 100 and 50 relapsed patients in the test and validation sets and the same number non-relapsed.

\begin{figure}[!htbp]
    \centering
\includegraphics[width=13.76cm,height=7.71cm]{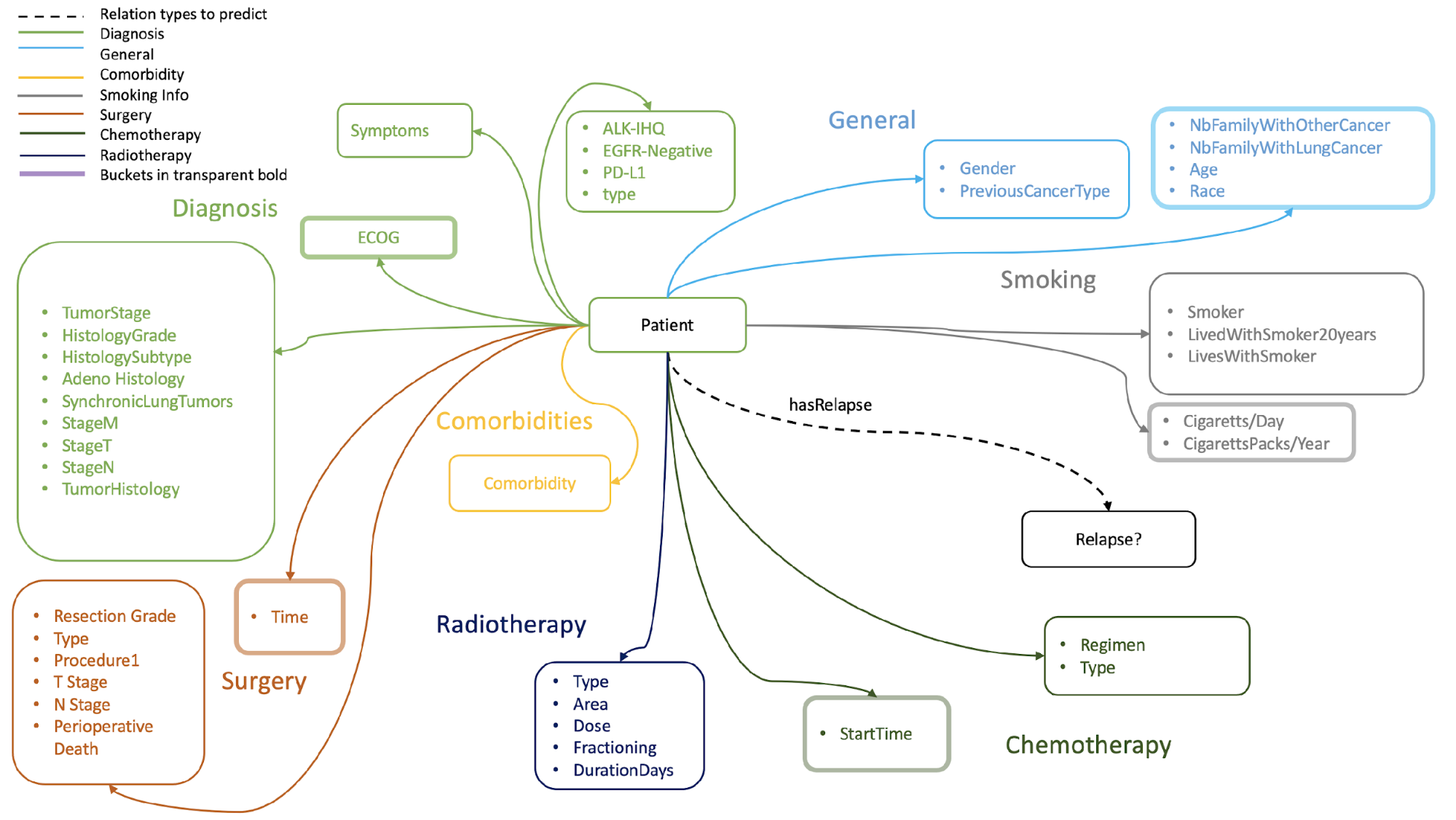}
    \caption{Diagram of the clinical data modelled as a knowledge graph.}
    \label{fig:my_label}
\end{figure}

\textbf{Model Selection, Hyperparameter values. }We adopt the ComplEx-N3 model (achieving state-of-the-art results \cite{lacroix_canonical_2018}). We select hyperparameter values of the KGE model  with a grid search across the following values: embedding size (50/100/200), number of negatives generated at runtime during training for each positive (5/25), and initializer (Xavier/uniform). For each parameter configuration (12), we select the best model according to accuracy score on the validation set. The best hyperparameter: Adam optimizer, learning rate$=$1e-3, multiclass-NLL loss, L3-regularizer with weight lambda$=$1e-3 training epochs$=$4000 with batches count$=$5, early-stopping. Embedding dimension is k$=$100, synthetic negatives ratio eta$=$25 and uniform initializer. Random seed set to default value.

\textbf{Implementation. }All experiments are implemented using Python3.7 with the Knowledge Graph Embedding library AmpliGraph \cite{costabello_accentureampligraph_2021} v1.4.0, using TensorFlow$=$1.15.2. We used Ubuntu 16.04 on an Intel Xeon Gold 6142, 64 GB, equipped with a Tesla V100 16GB.

\textbf{Evaluation Protocol: }

\textbf{\textit{Binary Classification}. }To compare with the tabular machine learning models, we evaluate graph models with binary classification metrics. For the best model, we choose the threshold for binary classification on the validation set using a value that maximizes difference between true positive  (TPR) and false positive (FPR) rates. We adopted this method over other approaches as it gives better validation results\footnote{ Other method include: selecting a threshold that maximizes the f1 score or selecting a label by assessment of which prediction was stronger (``does not recur" or ``recur").Both methods lead to worse results.}. 

\textbf{\textit{Ranking-based.} }We sanity-check the graph machine learning approach by evaluating using the learning-to-rank-based protocol and metrics commonly adopted in literature for these family of models. We use the following metrics: Hits@N, Mean Reciprocal Rank (MRR)\footnote{ For details see: Costabello, Luca et al. Knowledge Graph Embeddings Tutorial: From Theory to Practice, https://kge-tutorial-ecai2020.github.io/, ECAI’20. }. 

\textbf{Explanations. }For both tabular and graph machine learning, we adopt explainable AI approaches to provide context for the predictions. We adopt SHapley Additive exPlanations SHAP \cite{lundberg_unified_2017} for tabular models, and an example-based approach for graph machine learning. Along with the prediction score for relapse, models also return an accompanying chart or textual description supporting such prediction (Figure 3). Tabular models’ explanation is a waterfall chart with patient’s features contributing to the prediction (positively/negatively). For the graph machine learning model, we highlight similar past patients in the model space to the one being predicted (i.e. ``example-based" explanation).

\section{Results}

We report a patient’s characteristics of the selected 1,387 early-stage NSCLC patient cohort in Table 2.

\subsection{Patients’ Inclusion/Exclusion Criteria}

\begin{table}[!htbp]
\renewcommand{\arraystretch}{1.3}
\begin{adjustbox}{max width=\textwidth}
\begin{tabular}{p{4.52cm}p{3.62cm}p{3.47cm}p{5.71cm}p{4.52cm}p{3.62cm}p{3.47cm}p{5.71cm}}
\hline
\multicolumn{1}{|p{4.52cm}}{\textbf{Feature Name:}} & 
\multicolumn{1}{|p{3.62cm}}{\textbf{Case 1 \newline }\textit{Include a patient if the listed feature’s value is among one listed in this column.}} &
\multicolumn{1}{|p{3.47cm}}{\textbf{Case 2  \newline}\textit{Include a patient if the listed feature’s value is among one listed in this column.}} & 
\multicolumn{1}{|p{5.71cm}|}{\textbf{Exclude  \newline}\textit{Reference of which patients were excluded, because of feature’s values not being listed in Case 1 and/or Case 2.}} \\ 
\hline
\multicolumn{1}{|p{4.52cm}}{Cancer Type} & 
\multicolumn{2}{|p{7.09cm}}{Non-Small-Cell Lung Cancer (NSCLC)} & 
\multicolumn{1}{|p{5.71cm}|}{Small cell lung cancer, Carcinoid tumor, Epithelial thymoma, Mesothelioma, NA$\ast$} \\ 
\hline
\multicolumn{1}{|p{4.52cm}}{Tumor Stage} & 
\multicolumn{1}{|p{3.62cm}}{I, IA, IA1, IA2, IA3, IB, II, IIA, IIB} & 
\multicolumn{1}{|p{3.47cm}}{IIB} & 
\multicolumn{1}{|p{5.71cm}|}{IIIA, IIIB, IIIC, Otros, IV, IVA, IVB, Limited, Extended, NA$\ast$} \\ 
\hline
\multicolumn{1}{|p{4.52cm}}{Stage T} & 
\multicolumn{1}{|p{3.62cm}}{T1a, T1b, T1c, T2a, T2b, T0, Tx, Tis} & 
\multicolumn{1}{|p{3.47cm}}{T3} & 
\multicolumn{1}{|p{5.71cm}|}{T3 (except Case 2), T4, NA$\ast$} \\ 
\hline
\multicolumn{1}{|p{4.52cm}}{Stage N} & 
\multicolumn{1}{|p{3.62cm}}{N0, N1} & 
\multicolumn{1}{|p{3.47cm}}{N0} & 
\multicolumn{1}{|p{5.71cm}|}{N2, N3, Nx, NA$\ast$} \\ 
\hline
\multicolumn{1}{|p{4.52cm}}{Stage M} & 
\multicolumn{2}{|p{7.09cm}}{M0} & 
\multicolumn{1}{|p{3.47cm}|}{M1a, M1b, M1c, Mx, NA$\ast$} \\ 
\hline
\multicolumn{1}{|p{4.52cm}}{\textbf{\textit{Total number of patients }} \newline
\textbf{\textit{per case:}}} & 
\multicolumn{1}{|p{3.62cm}}{1711} & 
\multicolumn{1}{|p{3.47cm}}{375} & 
\multicolumn{1}{|p{5.71cm}|}{\multirow{2}{*}{\parbox{5.71cm}{-}}} \\ 
\hhline{---~}
\multicolumn{1}{|p{4.52cm}}{\textbf{\textit{Total number of patients }} \newline
\textbf{\textit{(Cases 1 $\&$ 2):}}} & 
\multicolumn{2}{|p{7.09cm}}{2086} & 
\multicolumn{1}{|p{3.47cm}|}{} \\ 
\hline
\multicolumn{1}{|p{4.52cm}}{Chemotherapy Type \newline
In the case when a patient had chemotherapy.} & 
\multicolumn{2}{|p{7.09cm}}{Adjuvant chemotherapy,  \newline
CHT intravenous,  \newline
Adjuvant CHT and RT, \newline
Neoadjuvant chemotherapy,  \newline
Neoadjuvant CHT-RT} & 
\multicolumn{1}{|p{5.71cm}|}{Oral targeted therapy, Oral and intravenous chemotherapy, Others, Immunotherapy, Concomitant CHT-RT, Sequential CHT-RT, Oral chemotherapy, Intravenous chemotherapy + immunotherapy, Hormonal, NA$\ast$} \\ 
\hline
\multicolumn{1}{|p{4.52cm}}{Include patients who did not undergo chemotherapy.} & 
\multicolumn{2}{|p{7.09cm}}{\textit{Include patients without chemotherapy that fit other criteria.}} & 
\multicolumn{1}{|p{3.47cm}|}{-} \\ 
\hline
\multicolumn{1}{|p{4.52cm}}{Radiotherapy Intention \newline
\textit{In the case when a patient had radiotherapy.}} & 
\multicolumn{2}{|p{7.09cm}}{Adjuvant} & 
\multicolumn{1}{|p{5.71cm}|}{Palliative, Radical, Prophylactic, Neoadjuvant, NA$\ast$} \\ 
\hline
\multicolumn{1}{|p{4.52cm}}{Include patients who did not undergo radiotherapy.} & 
\multicolumn{2}{|p{7.09cm}}{\textit{Include patients without radiotherapy that fit other criteria.}} & 
\multicolumn{1}{|p{3.47cm}|}{\multirow{2}{*}{\parbox{3.47cm}{-}}} \\ 
\hhline{---~}
\multicolumn{1}{|p{4.52cm}}{\textbf{\textit{Full Cohort Size:}}} & 
\multicolumn{2}{|p{7.09cm}}{\textbf{1387}} & 
\multicolumn{1}{|p{3.47cm}|}{} \\ 
\hline
\end{tabular}
\end{adjustbox}
\newline
\newline
\textbf{$\ast$NA - Not Available meaning there is entry about a patient, but value is either NULL or - or NA.}
\newline
\newline

\caption{Patients Inclusion/Exclusion Criteria for training ML models. For the task of prediction of relapse or progression for early-stage lung cancer patients, patients were selected with the inclusion/exclusion criteria listed in this table.}
\end{table}

\subsection{Patients’ Characteristics}
\begin{table}[H]
\renewcommand{\arraystretch}{1.4}
\begin{adjustbox}{max width=\textwidth}
\begin{tabular}{p{3.5cm}p{3.76cm}p{2.49cm}p{2.78cm}p{4.18cm}p{2.62cm}p{3.76cm}p{2.49cm}p{2.78cm}p{4cm}}
\hline
\multicolumn{2}{|p{6.38cm}}{\multirow{2}{*}{\parbox{6.38cm}{\textbf{Characteristic}}}} & 
\multicolumn{1}{|p{3.76cm}}{\textbf{\uline{Relapse}}} & 
\multicolumn{1}{|p{2.49cm}}{\textbf{\uline{No relapse}}} & 
\multicolumn{1}{|p{2.78cm}|}{\textbf{\uline{Total}}} \\ 
\hhline{~~---}
\multicolumn{2}{|p{6.38cm}}{} & 
\multicolumn{1}{|p{3.76cm}}{\textbf{517 (37.3$\%$)}} & 
\multicolumn{1}{|p{2.49cm}}{\textbf{870 (62.7$\%$)}} & 
\multicolumn{1}{|p{2.78cm}|}{\textbf{1387 (100.0$\%$)}} \\ 
\hline
\multicolumn{1}{|p{2.62cm}}{Age} & 
\multicolumn{1}{|p{3.76cm}}{Mean (range)} & 
\multicolumn{1}{|p{2.49cm}}{65.7 (33-88)} & 
\multicolumn{1}{|p{2.78cm}}{65.6 (31-118)} & 
\multicolumn{1}{|p{4.18cm}|}{65.7 (31-118)} \\ 
\hline
\multicolumn{1}{|p{2.62cm}}{\multirow{2}{*}{\parbox{2.62cm}{Gender}}} & 
\multicolumn{1}{|p{3.76cm}}{Male} & 
\multicolumn{1}{|p{2.49cm}}{407 (39.0$\%$)} & 
\multicolumn{1}{|p{2.78cm}}{636 (61.0$\%$)} & 
\multicolumn{1}{|p{4.18cm}|}{1043 (75.2$\%$)} \\ 
\hhline{~----}
\multicolumn{1}{|p{2.62cm}}{} & 
\multicolumn{1}{|p{3.76cm}}{Female} & 
\multicolumn{1}{|p{2.49cm}}{110 (32.0$\%$)} & 
\multicolumn{1}{|p{2.78cm}}{234 (68.0$\%$)} & 
\multicolumn{1}{|p{4.18cm}|}{344 (24.8$\%$)} \\ 
\hline
\multicolumn{1}{|p{2.62cm}}{\multirow{2}{*}{\parbox{2.62cm}{Smoking history}}} & 
\multicolumn{1}{|p{3.76cm}}{Current/Previous} & 
\multicolumn{1}{|p{2.49cm}}{460 (38.0$\%$)} & 
\multicolumn{1}{|p{2.78cm}}{750 (62.0$\%$)} & 
\multicolumn{1}{|p{4.18cm}|}{1210 (87.2$\%$)} \\ 
\hhline{~----}
\multicolumn{1}{|p{2.62cm}}{} & 
\multicolumn{1}{|p{3.76cm}}{Non-Smoker} & 
\multicolumn{1}{|p{2.49cm}}{57 (32.2$\%$)} & 
\multicolumn{1}{|p{2.78cm}}{120 (68.8$\%$)} & 
\multicolumn{1}{|p{4.18cm}|}{177 (12.8$\%$)} \\ 
\hline
\multicolumn{1}{|p{2.62cm}}{\multirow{8}{*}{\parbox{2.62cm}{Cancer stage}}} & 
\multicolumn{1}{|p{3.76cm}}{I} & 
\multicolumn{1}{|p{2.49cm}}{1 (100.0$\%$)} & 
\multicolumn{1}{|p{2.78cm}}{0 (0.0$\%$)} & 
\multicolumn{1}{|p{4.18cm}|}{1 (0.0721$\%$)} \\ 
\hhline{~----}
\multicolumn{1}{|p{2.62cm}}{} & 
\multicolumn{1}{|p{3.76cm}}{IA1} & 
\multicolumn{1}{|p{2.49cm}}{8 (32.0$\%$)} & 
\multicolumn{1}{|p{2.78cm}}{17 (68.0$\%$)} & 
\multicolumn{1}{|p{4.18cm}|}{25 (1.8$\%$)} \\ 
\hhline{~----}
\multicolumn{1}{|p{2.62cm}}{} & 
\multicolumn{1}{|p{3.76cm}}{IA3} & 
\multicolumn{1}{|p{2.49cm}}{8 (17.4$\%$)} & 
\multicolumn{1}{|p{2.78cm}}{38 (82.6$\%$)} & 
\multicolumn{1}{|p{4.18cm}|}{46 (3.32$\%$)} \\ 
\hhline{~----}
\multicolumn{1}{|p{2.62cm}}{} & 
\multicolumn{1}{|p{3.76cm}}{IA2} & 
\multicolumn{1}{|p{2.49cm}}{9 (13.6$\%$)} & 
\multicolumn{1}{|p{2.78cm}}{57 (86.4$\%$)} & 
\multicolumn{1}{|p{4.18cm}|}{66 (4.76$\%$)} \\ 
\hhline{~----}
\multicolumn{1}{|p{2.62cm}}{} & 
\multicolumn{1}{|p{3.76cm}}{IIA} & 
\multicolumn{1}{|p{2.49cm}}{112 (46.7$\%$)} & 
\multicolumn{1}{|p{2.78cm}}{128 (53.3$\%$)} & 
\multicolumn{1}{|p{4.18cm}|}{240 (17.3$\%$)} \\ 
\hhline{~----}
\multicolumn{1}{|p{2.62cm}}{} & 
\multicolumn{1}{|p{3.76cm}}{IA} & 
\multicolumn{1}{|p{2.49cm}}{74 (29.0$\%$)} & 
\multicolumn{1}{|p{2.78cm}}{181 (71.0$\%$)} & 
\multicolumn{1}{|p{4.18cm}|}{255 (18.4$\%$)} \\ 
\hhline{~----}
\multicolumn{1}{|p{2.62cm}}{} & 
\multicolumn{1}{|p{3.76cm}}{IIB} & 
\multicolumn{1}{|p{2.49cm}}{148 (41.9$\%$)} & 
\multicolumn{1}{|p{2.78cm}}{205 (58.1$\%$)} & 
\multicolumn{1}{|p{4.18cm}|}{353 (25.5$\%$)} \\ 
\hhline{~----}
\multicolumn{1}{|p{2.62cm}}{} & 
\multicolumn{1}{|p{3.76cm}}{IB} & 
\multicolumn{1}{|p{2.49cm}}{157 (39.2$\%$)} & 
\multicolumn{1}{|p{2.78cm}}{244 (60.8$\%$)} & 
\multicolumn{1}{|p{4.18cm}|}{401 (28.9$\%$)} \\ 
\hline
\multicolumn{1}{|p{2.62cm}}{\multirow{7}{*}{\parbox{2.62cm}{T stage}}} & 
\multicolumn{1}{|p{3.76cm}}{T2a} & 
\multicolumn{1}{|p{2.49cm}}{210 (41.1$\%$)} & 
\multicolumn{1}{|p{2.78cm}}{301 (58.9$\%$)} & 
\multicolumn{1}{|p{4.18cm}|}{511 (36.8$\%$)} \\ 
\hhline{~----}
\multicolumn{1}{|p{2.62cm}}{} & 
\multicolumn{1}{|p{3.76cm}}{T1b} & 
\multicolumn{1}{|p{2.49cm}}{56 (26.0$\%$)} & 
\multicolumn{1}{|p{2.78cm}}{159 (74.0$\%$)} & 
\multicolumn{1}{|p{4.18cm}|}{215 (15.5$\%$)} \\ 
\hhline{~----}
\multicolumn{1}{|p{2.62cm}}{} & 
\multicolumn{1}{|p{3.76cm}}{T3} & 
\multicolumn{1}{|p{2.49cm}}{94 (42.0$\%$)} & 
\multicolumn{1}{|p{2.78cm}}{130 (68.0$\%$)} & 
\multicolumn{1}{|p{4.18cm}|}{224 (16.1$\%$)} \\ 
\hhline{~----}
\multicolumn{1}{|p{2.62cm}}{} & 
\multicolumn{1}{|p{3.76cm}}{T2b} & 
\multicolumn{1}{|p{2.49cm}}{76 (42.5$\%$)} & 
\multicolumn{1}{|p{2.78cm}}{103 (57.5$\%$)} & 
\multicolumn{1}{|p{4.18cm}|}{179 (12.9$\%$)} \\ 
\hhline{~----}
\multicolumn{1}{|p{2.62cm}}{} & 
\multicolumn{1}{|p{3.76cm}}{T1a} & 
\multicolumn{1}{|p{2.49cm}}{55 (32.7$\%$)} & 
\multicolumn{1}{|p{2.78cm}}{113 (67.3$\%$)} & 
\multicolumn{1}{|p{4.18cm}|}{168 (12.1$\%$)} \\ 
\hhline{~----}
\multicolumn{1}{|p{2.62cm}}{} & 
\multicolumn{1}{|p{3.76cm}}{T1c} & 
\multicolumn{1}{|p{2.49cm}}{14 (20.0$\%$)} & 
\multicolumn{1}{|p{2.78cm}}{56 (80.0$\%$)} & 
\multicolumn{1}{|p{4.18cm}|}{70 (5.05$\%$)} \\ 
\hhline{~----}
\multicolumn{1}{|p{2.62cm}}{} & 
\multicolumn{1}{|p{3.76cm}}{Tx} & 
\multicolumn{1}{|p{2.49cm}}{12 (60.0$\%$)} & 
\multicolumn{1}{|p{2.78cm}}{8 (40.0$\%$)} & 
\multicolumn{1}{|p{4.18cm}|}{20 (1.44$\%$)} \\ 
\hline
\multicolumn{1}{|p{2.62cm}}{\multirow{2}{*}{\parbox{2.62cm}{N stage}}} & 
\multicolumn{1}{|p{3.76cm}}{N0} & 
\multicolumn{1}{|p{2.49cm}}{406 (35.2$\%$)} & 
\multicolumn{1}{|p{2.78cm}}{747 (64.8$\%$)} & 
\multicolumn{1}{|p{4.18cm}|}{1153 (83.1$\%$)} \\ 
\hhline{~----}
\multicolumn{1}{|p{2.62cm}}{} & 
\multicolumn{1}{|p{3.76cm}}{N1} & 
\multicolumn{1}{|p{2.49cm}}{111 (47.4$\%$)} & 
\multicolumn{1}{|p{2.78cm}}{123 (52.6$\%$)} & 
\multicolumn{1}{|p{4.18cm}|}{234 (16.9$\%$)} \\ 
\hline
\multicolumn{1}{|p{2.62cm}}{M stage} & 
\multicolumn{1}{|p{3.76cm}}{M0} & 
\multicolumn{1}{|p{2.49cm}}{517 (37.3$\%$)} & 
\multicolumn{1}{|p{2.78cm}}{870 (62.7$\%$)} & 
\multicolumn{1}{|p{4.18cm}|}{1387 (100.0$\%$)} \\ 
\hline
\multicolumn{1}{|p{2.62cm}}{\multirow{5}{*}{\parbox{2.62cm}{ECOG status}}} & 
\multicolumn{1}{|p{3.76cm}}{0} & 
\multicolumn{1}{|p{2.49cm}}{259 (30.3$\%$)} & 
\multicolumn{1}{|p{2.78cm}}{597 (69.7$\%$)} & 
\multicolumn{1}{|p{4.18cm}|}{856 (61.7$\%$)} \\ 
\hhline{~----}
\multicolumn{1}{|p{2.62cm}}{} & 
\multicolumn{1}{|p{3.76cm}}{1} & 
\multicolumn{1}{|p{2.49cm}}{225 (47.3$\%$)} & 
\multicolumn{1}{|p{2.78cm}}{251 (52.7$\%$)} & 
\multicolumn{1}{|p{4.18cm}|}{476 (34.3$\%$)} \\ 
\hhline{~----}
\multicolumn{1}{|p{2.62cm}}{} & 
\multicolumn{1}{|p{3.76cm}}{2} & 
\multicolumn{1}{|p{2.49cm}}{26 (60.5$\%$)} & 
\multicolumn{1}{|p{2.78cm}}{17 (39.5$\%$)} & 
\multicolumn{1}{|p{4.18cm}|}{43 (3.1$\%$)} \\ 
\hhline{~----}
\multicolumn{1}{|p{2.62cm}}{} & 
\multicolumn{1}{|p{3.76cm}}{3} & 
\multicolumn{1}{|p{2.49cm}}{6 (75.0$\%$)} & 
\multicolumn{1}{|p{2.78cm}}{2 (25.0$\%$)} & 
\multicolumn{1}{|p{4.18cm}|}{8 (0.577$\%$)} \\ 
\hhline{~----}
\multicolumn{1}{|p{2.62cm}}{} & 
\multicolumn{1}{|p{3.76cm}}{4} & 
\multicolumn{1}{|p{2.49cm}}{1 (100.0$\%$)} & 
\multicolumn{1}{|p{2.78cm}}{0 (0.0$\%$)} & 
\multicolumn{1}{|p{4.18cm}|}{1 (0.0721$\%$)} \\ 
\hline
\multicolumn{1}{|p{3.5cm}}{\multirow{5}{*}{\parbox{3.5cm}{Tumor Differentiation}}} & 
\multicolumn{1}{|p{3.76cm}}{Non specified} & 
\multicolumn{1}{|p{2.49cm}}{146 (40.3$\%$)} & 
\multicolumn{1}{|p{2.78cm}}{216 (69.7$\%$)} & 
\multicolumn{1}{|p{4.18cm}|}{362 (26.1$\%$)} \\ 
\hhline{~----}
\multicolumn{1}{|p{2.62cm}}{} & 
\multicolumn{1}{|p{3.76cm}}{Moderately differentiated} & 
\multicolumn{1}{|p{2.49cm}}{56 (27.1$\%$)} & 
\multicolumn{1}{|p{2.78cm}}{151 (72.9$\%$)} & 
\multicolumn{1}{|p{4.18cm}|}{207 (14.9$\%$)} \\ 
\hhline{~----}
\multicolumn{1}{|p{2.62cm}}{} & 
\multicolumn{1}{|p{3.76cm}}{Poorly differentiated} & 
\multicolumn{1}{|p{2.49cm}}{45 (43.3$\%$)} & 
\multicolumn{1}{|p{2.78cm}}{59 (56.7$\%$)} & 
\multicolumn{1}{|p{4.18cm}|}{104 (7.5$\%$)} \\ 
\hhline{~----}
\multicolumn{1}{|p{2.62cm}}{} & 
\multicolumn{1}{|p{3.76cm}}{Well differentiated} & 
\multicolumn{1}{|p{2.49cm}}{56 (43.3$\%$)} & 
\multicolumn{1}{|p{2.78cm}}{74 (56.9$\%$)} & 
\multicolumn{1}{|p{4.18cm}|}{130 (9.37$\%$)} \\ 
\hhline{~----}
\multicolumn{1}{|p{2.62cm}}{} & 
\multicolumn{1}{|p{3.76cm}}{Undifferentiated} & 
\multicolumn{1}{|p{2.49cm}}{3 (50.0$\%$)} & 
\multicolumn{1}{|p{2.78cm}}{3 (50.0$\%$)} & 
\multicolumn{1}{|p{4.18cm}|}{6 (0.433$\%$)} \\ 
\hline
\multicolumn{2}{|p{6.38cm}}{Surgery} & 
\multicolumn{1}{|p{3.76cm}}{459 (33.1$\%$)} & 
\multicolumn{1}{|p{2.49cm}}{848 (61.1$\%$)} & 
\multicolumn{1}{|p{2.78cm}|}{1307 (94.2$\%$)} \\ 
\hline
\multicolumn{2}{|p{6.38cm}}{Chemotherapy} & 
\multicolumn{1}{|p{3.76cm}}{379 (27.3$\%$)} & 
\multicolumn{1}{|p{2.49cm}}{318 (22.9$\%$)} & 
\multicolumn{1}{|p{2.78cm}|}{697 (50.3$\%$)} \\ 
\hline
\multicolumn{2}{|p{6.38cm}}{Radiotherapy} & 
\multicolumn{1}{|p{3.76cm}}{42 (3.03$\%$)} & 
\multicolumn{1}{|p{2.49cm}}{22 (1.59$\%$)} & 
\multicolumn{1}{|p{2.78cm}|}{64 (4.61$\%$)} \\ 
\hline
\end{tabular}
\end{adjustbox}
\caption{Cohort analysis of the early-stage patients, which were utilized to build the dataset (for full list of attributes used for training the models see Supplementary Material. Inclusion and exclusion criteria are described in the section above). Compare with Lung cancer in Spain: information from the Thoracic Tumors Registry (TTR study) for a set of NSCLC patients characteristics (Provencio et al. 2019 \cite{provencio_lung_2019}) and \cite{sameh_k_mohamed_et_al_predicting_2021} for a cohort of all NSCLC patients in the dataset.}
\end{table}

\subsection{Predictive Power Assessment}
The goal is to categorize patients into individuals likely to relapse or not (regardless of when the relapse occurs). The problem is cast as a binary classification task to predict the probability of relapse and support it with explanations. We cannot directly compare against baseline models from our previous work \cite{sameh_k_mohamed_et_al_predicting_2021} due to different inclusion criteria for patients, resulting in smaller training dataset; we report a random baseline instead. For both types of models tabular (Logistic Regression, Random Forest, Gradient Boosting) and graph model (ComplEx-N3) we run hyperparameter search experiments to pick best parameters on the validation set and then we run the ablation study with different group of features 1) diagnosis only features, 2) diagnosis and surgery and 3) diagnosis, surgery, and treatment, presented in Table 3. To evaluate the task, we use a set of binary metrics for two types of models, reported in Tables 3. See supplementary materials for metrics details.

\textbf{Machine Learning Predictive Performance Results}

\begin{table}[!htbp]
\renewcommand{\arraystretch}{1.3}
\begin{adjustbox}{max width=\textwidth}
\begin{tabular}{p{2.15cm}p{2.01cm}p{2.01cm}p{2.01cm}p{2.01cm}p{2.01cm}p{2.01cm}p{2.5cm}p{2.15cm}p{2.01cm}p{2.01cm}p{2.01cm}p{2.01cm}p{2.01cm}p{2.01cm}p{2.5cm}}
\hhline{~~~~~~~~}
\multicolumn{1}{p{2.15cm}}{} & 
\multicolumn{1}{p{2.01cm}}{\textbf{Model}} & 
\multicolumn{1}{|p{2.01cm}}{\textbf{Accuracy}} & 
\multicolumn{1}{p{2.01cm}}{\textbf{Precision}} & 
\multicolumn{1}{p{2.01cm}}{\textbf{Recall}} & 
\multicolumn{1}{p{2.01cm}}{\textbf{F1}} & 
\multicolumn{1}{p{2.01cm}}{\textbf{AUC-PR\footnote{ Also known as average precision/ AP/ area under precision-recall curve, to not confuse with the mean value of precision (Precision column).}}} & 
\multicolumn{1}{p{2.5cm}}{\textbf{AUC-ROC}} \\ 
\hline
\multicolumn{1}{p{2.15cm}}{} & 
\multicolumn{7}{p{14.559999999999999cm}|}{\textbf{Diagnosis Features}} \\ 
\hhline{~-~~~~~~}
\multicolumn{1}{p{2.15cm}}{Tabular} & 
\multicolumn{1}{p{2.01cm}}{Random Baseline} & 
\multicolumn{1}{p{2.01cm}}{0.505$\pm$0.040} & 
\multicolumn{1}{p{2.01cm}}{0.378$\pm$0.041} & 
\multicolumn{1}{p{2.01cm}}{0.511$\pm$0.061} & 
\multicolumn{1}{p{2.01cm}}{0.434$\pm$0.047} & 
\multicolumn{1}{p{2.01cm}}{0.373$\pm$0.002} & 
\multicolumn{1}{p{2.5cm}}{0.500$\pm$0.000} \\ 
\hhline{~~~~~~~~}
\multicolumn{1}{p{2.15cm}}{} & 
\multicolumn{1}{p{2.01cm}}{Gradient Boosting} & 
\multicolumn{1}{p{2.01cm}}{0.636$\pm$0.040} & 
\multicolumn{1}{p{2.01cm}}{0.511$\pm$0.051} & 
\multicolumn{1}{p{2.01cm}}{0.540$\pm$0.083} & 
\multicolumn{1}{p{2.01cm}}{0.523$\pm$0.062} & 
\multicolumn{1}{p{2.01cm}}{0.573$\pm$0.069} & 
\multicolumn{1}{p{2.5cm}}{0.674$\pm$0.045} \\ 
\hhline{~~~~~~~~}
\multicolumn{1}{p{2.15cm}}{} & 
\multicolumn{1}{p{2.01cm}}{Logistic Regression} & 
\multicolumn{1}{p{2.01cm}}{0.613$\pm$0.026} & 
\multicolumn{1}{p{2.01cm}}{0.485$\pm$0.027} & 
\multicolumn{1}{p{2.01cm}}{\textbf{0.577$\pm$0.052}} & 
\multicolumn{1}{p{2.01cm}}{\textbf{0.526$\pm$0.030}} & 
\multicolumn{1}{p{2.01cm}}{0.575$\pm$0.053} & 
\multicolumn{1}{p{2.5cm}}{\textbf{0.675$\pm$0.037}} \\ 
\hhline{~~~~~~~~}
\multicolumn{1}{p{2.15cm}}{} & 
\multicolumn{1}{p{2.01cm}}{Random Forest} & 
\multicolumn{1}{p{2.01cm}}{\textbf{0.663$\pm$0.031}} & 
\multicolumn{1}{p{2.01cm}}{\textbf{0.567$\pm$0.057}} & 
\multicolumn{1}{p{2.01cm}}{0.418$\pm$0.060} & 
\multicolumn{1}{p{2.01cm}}{0.480$\pm$0.053} & 
\multicolumn{1}{p{2.01cm}}{\textbf{0.576$\pm$0.050}} & 
\multicolumn{1}{p{2.5cm}}{0.670$\pm$0.031} \\ 
\hhline{~~~~~~~~}
\multicolumn{1}{p{2.15cm}}{Graph} & 
\multicolumn{1}{p{2.01cm}}{ComplEx-N3} & 
\multicolumn{1}{|p{2.01cm}}{0.44} & 
\multicolumn{1}{p{2.01cm}}{0.4362} & 
\multicolumn{1}{p{2.01cm}}{0.41} & 
\multicolumn{1}{p{2.01cm}}{0.\textbf{4227}} & 
\multicolumn{1}{p{2.01cm}}{0.4738} & 
\multicolumn{1}{p{2.5cm}}{0.44} \\ 
\hline
\multicolumn{1}{p{2.15cm}}{} & 
\multicolumn{7}{p{14.559999999999999cm}|}{\textbf{Diag. Feat. + Surgery}} \\ 
\hhline{--~~~~~~}
\multicolumn{1}{p{2.15cm}}{Tabular} & 
\multicolumn{1}{p{2.01cm}}{Random Baseline} & 
\multicolumn{1}{p{2.01cm}}{0.481$\pm$0.032} & 
\multicolumn{1}{p{2.01cm}}{0.349$\pm$0.038} & 
\multicolumn{1}{p{2.01cm}}{0.458$\pm$0.080} & 
\multicolumn{1}{p{2.01cm}}{0.395$\pm$0.052} & 
\multicolumn{1}{p{2.01cm}}{0.373$\pm$0.002} & 
\multicolumn{1}{p{2.5cm}}{0.500$\pm$0.000} \\ 
\hhline{~~~~~~~~}
\multicolumn{1}{p{2.15cm}}{} & 
\multicolumn{1}{p{2.01cm}}{Gradient Boosting} & 
\multicolumn{1}{p{2.01cm}}{0.687$\pm$0.025} & 
\multicolumn{1}{p{2.01cm}}{0.589$\pm$0.042} & 
\multicolumn{1}{p{2.01cm}}{0.549$\pm$0.020} & 
\multicolumn{1}{p{2.01cm}}{0.567$\pm$0.020} & 
\multicolumn{1}{p{2.01cm}}{0.637$\pm$0.046} & 
\multicolumn{1}{p{2.5cm}}{0.721$\pm$0.030} \\ 
\hhline{~~~~~~~~}
\multicolumn{1}{p{2.15cm}}{} & 
\multicolumn{1}{p{2.01cm}}{Logistic Regression} & 
\multicolumn{1}{p{2.01cm}}{0.663$\pm$0.027} & 
\multicolumn{1}{p{2.01cm}}{0.546$\pm$0.035} & 
\multicolumn{1}{p{2.01cm}}{0.582$\pm$0.031} & 
\multicolumn{1}{p{2.01cm}}{0.563$\pm$0.027} & 
\multicolumn{1}{p{2.01cm}}{0.616$\pm$0.057} & 
\multicolumn{1}{p{2.5cm}}{0.711$\pm$0.035} \\ 
\hhline{~~~~~~~~}
\multicolumn{1}{p{2.15cm}}{} & 
\multicolumn{1}{p{2.01cm}}{Random Forest} & 
\multicolumn{1}{p{2.01cm}}{\textbf{0.707$\pm$0.038}} & 
\multicolumn{1}{p{2.01cm}}{0.647$\pm$0.073} & 
\multicolumn{1}{p{2.01cm}}{0.482$\pm$0.056} & 
\multicolumn{1}{p{2.01cm}}{0.550$\pm$0.053} & 
\multicolumn{1}{p{2.01cm}}{\textbf{0.645$\pm$0.060}} & 
\multicolumn{1}{p{2.5cm}}{\textbf{0.728$\pm$0.033}} \\ 
\hhline{~~~~~~~~}
\multicolumn{1}{p{2.15cm}}{Graph} & 
\multicolumn{1}{p{2.01cm}}{ComplEx-N3} & 
\multicolumn{1}{|p{2.01cm}}{0.68} & 
\multicolumn{1}{p{2.01cm}}{0.6607} & 
\multicolumn{1}{p{2.01cm}}{\textbf{0.74}} & 
\multicolumn{1}{p{2.01cm}}{\textbf{0.6981}} & 
\multicolumn{1}{p{2.01cm}}{0.6189} & 
\multicolumn{1}{p{2.5cm}}{0.68} \\ 
\hline
\multicolumn{1}{p{2.15cm}}{} & 
\multicolumn{7}{p{14.559999999999999cm}|}{\textbf{Diag. Feat. + Surg. + Treatment}} \\ 
\hhline{--~~~~~~}
\multicolumn{1}{p{2.15cm}}{Tabular} & 
\multicolumn{1}{p{2.01cm}}{Random Baseline} & 
\multicolumn{1}{p{2.01cm}}{0.512$\pm$0.031} & 
\multicolumn{1}{p{2.01cm}}{0.384$\pm$0.028} & 
\multicolumn{1}{p{2.01cm}}{0.507$\pm$0.043} & 
\multicolumn{1}{p{2.01cm}}{0.436$\pm$0.030} & 
\multicolumn{1}{p{2.01cm}}{0.373$\pm$0.002} & 
\multicolumn{1}{p{2.5cm}}{0.500$\pm$0.000} \\ 
\hhline{~~~~~~~~}
\multicolumn{1}{p{2.15cm}}{} & 
\multicolumn{1}{p{2.01cm}}{Gradient Boosting} & 
\multicolumn{1}{p{2.01cm}}{0.746$\pm$0.026} & 
\multicolumn{1}{p{2.01cm}}{0.667$\pm$0.052} & 
\multicolumn{1}{p{2.01cm}}{0.649$\pm$0.088} & 
\multicolumn{1}{p{2.01cm}}{0.653$\pm$0.044} & 
\multicolumn{1}{p{2.01cm}}{0.765$\pm$0.032} & 
\multicolumn{1}{p{2.5cm}}{0.807$\pm$0.035} \\ 
\hhline{~~~~~~~~}
\multicolumn{1}{p{2.15cm}}{} & 
\multicolumn{1}{p{2.01cm}}{Logistic Regression} & 
\multicolumn{1}{p{2.01cm}}{0.735$\pm$0.039} & 
\multicolumn{1}{p{2.01cm}}{0.632$\pm$0.049} & 
\multicolumn{1}{p{2.01cm}}{\textbf{0.692$\pm$0.074}} & 
\multicolumn{1}{p{2.01cm}}{0.660$\pm$0.054} & 
\multicolumn{1}{p{2.01cm}}{0.708$\pm$0.052} & 
\multicolumn{1}{p{2.5cm}}{0.787$\pm$0.041} \\ 
\hhline{~~~~~~~~}
\multicolumn{1}{p{2.15cm}}{} & 
\multicolumn{1}{p{2.01cm}}{Random Forest} & 
\multicolumn{1}{p{2.01cm}}{\textbf{0.761$\pm$0.030}} & 
\multicolumn{1}{p{2.01cm}}{0.701$\pm$0.049} & 
\multicolumn{1}{p{2.01cm}}{0.632$\pm$0.074} & 
\multicolumn{1}{p{2.01cm}}{\textbf{0.662$\pm$0.049}} & 
\multicolumn{1}{p{2.01cm}}{0.762$\pm$0.035} & 
\multicolumn{1}{p{2.5cm}}{\textbf{0.813$\pm$0.031}} \\ 
\hhline{~~~~~~~~}
\multicolumn{1}{p{2.15cm}}{Graph ML} & 
\multicolumn{1}{p{2.01cm}}{ComplEx-N3} & 
\multicolumn{1}{|p{2.01cm}}{0.685} & 
\multicolumn{1}{p{2.01cm}}{\textbf{0.7176}} & 
\multicolumn{1}{p{2.01cm}}{0.61} & 
\multicolumn{1}{p{2.01cm}}{0.6594} & 
\multicolumn{1}{p{2.01cm}}{0.6327} & 
\multicolumn{1}{p{2.5cm}}{0.685} \\ 
\hhline{~~~~~~~~}
\end{tabular}
\end{adjustbox}
\newline
\newline
\caption{Binary Classification evaluation results for 3 tabular models (Gradient Boosting, Logistic Regression, Random Forest), 1 relational model (ComlEx-N3), and a baseline, all trained with three feature configurations.}
\end{table}

\subsection{Explanations, Examples}

We predict the probability of relapse of a patient using tabular and graph machine learning and report AI-generated explanations for each approach.

\textbf{Tabular model: }we adopt SHAP, a method based on Shapley Values to estimate the marginal contribution\textit{ }of each dimension on the predicted result. Shapley values is a game theory method that allocates the ``dividend" of the prediction to each dimension. The idea underneath SHAP is that dimensions of the target data point being predicted by the model compete against each other to contribute to the prediction. It uses a sample of training examples, to compute the expected output of the model. With waterfall plots present contribution of each feature towards the output. In the example below, we have in red positive and in blue negative contributions. Gradient Boosting Classifier predicts that the patient has 0.589 chances of  relapsing. Having four family members with other cancer is the highest contributing factor, which  increases the prediction by 0.19 . Missing information on the regimen of the chemotherapy or not having it done at all (chemotherapy@t1\_regimen\_$\#$nan$\#$) decreases the prediction by 0.07. This method has its limitations being it explains model prediction and not causal factors that may be at the root of the relapse. E.g., not providing information about the type of chemotherapy regimen does not increase the risk of relapse. 

\textbf{Graph Machine Learning model:} we adopt an example-based approach that provides influential examples the model is trained on. Figure 3 A) shows prediction for a patient of 0.31 probability of relapse. Three influential patients are returned as a result based on the learned latent space of the model. For one influential example (906926) common and differing characteristics are listed. In common are stage Ib, and TNM of:T2a, N0, M0, adenocarcinoma histology, and grade being moderately differentiated andECOG$=$0. Both patients are in the same age group: 60-65 and do not have family members with lung cancer. They Have Dyslipidemia, HTA and other comorbidities and are both smokers with one smoking roughly 5-10 cigarettes/day, both are asymptomatic. They went through the curative surgery in the first 10 months from diagnosis with resection grade R0 and the same TN stages as at the time of diagnosis. Despite these common traits they differ in the following characteristics: one has histology subtype Papillary and the other Acinar, they are of different genders, and one has family history of other cancer (3-5) while the other has none. Influential patients had Colorectal cancer, COPD and Hypercholesterolemia in the past while predicted patients had cardiopathy. In terms of the surgery both patients went through different procedures right inferior lobectomy and typical segmentectomy respectively.  This method limitation is that it does not rank the features, instead it brings historical cases that display similarity to the predicted patient according to trained model space. We can see that the method  brings to attention past cases of patients similar to the predicted patient, which  depends on the definition of similarity. We can augment methods to list more relevant cases by incorporating doctor-defined constraints e.g. both patients should have the same TNM stages and same procedure.

\begin{figure}[!htbp]
  {\centering  \textbf{A) Tabular model}\ \ \ \ \ \ \ \ \ \ \ }\par  \ \ \ \ \ \ \ \ 
 {   \centering
\subcaptionbox{Tabular model with 75$\%$ accuracy, trained over 1,348 patients. SHAP explanation with a waterfall plot of features contributing to the prediction, in red increasing the prediction score, in blue decreasing.}{\includegraphics[width=\textwidth]{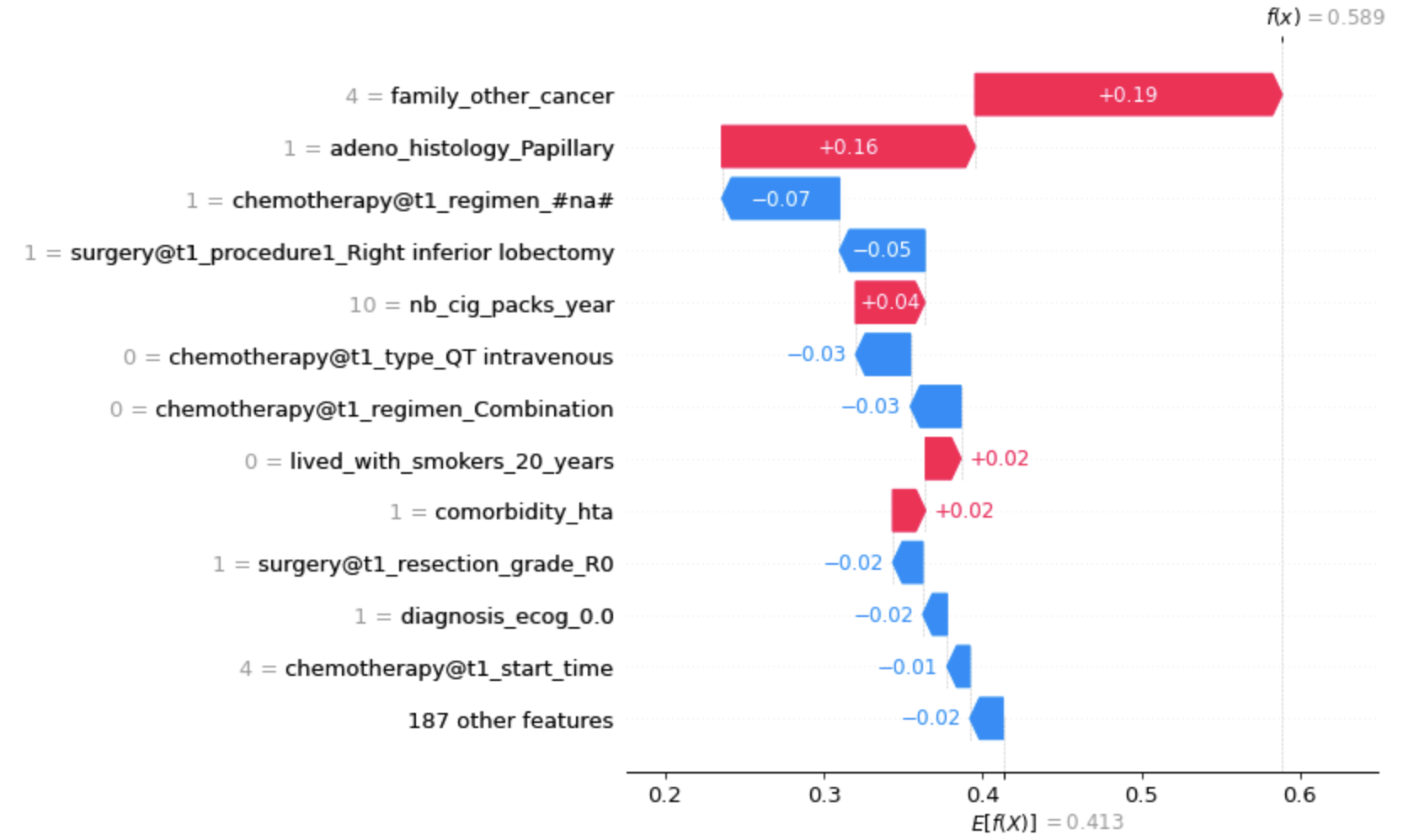}}}
{\textbf{B)} \textbf{Graph Machine Learning Model }\ \ \ \ }\par
\hfil
\subcaptionbox{Graph Machine Learning model with 68$\%$ accuracy, trained over 1,348 patients. Example-based explanation. 1) Prediction Summary for selected patients including predicted risk, number of similar examples, and number of training cases. 2) Retrieved Exemplary Cases, i.e., ‘influential patients’. 3) Commonalities and Differences between the patient being predicted and the selected influential patient retrieved by the example-based explanation method. a) Venn diagram view b) Table view.}{
\includegraphics[width=\textwidth]{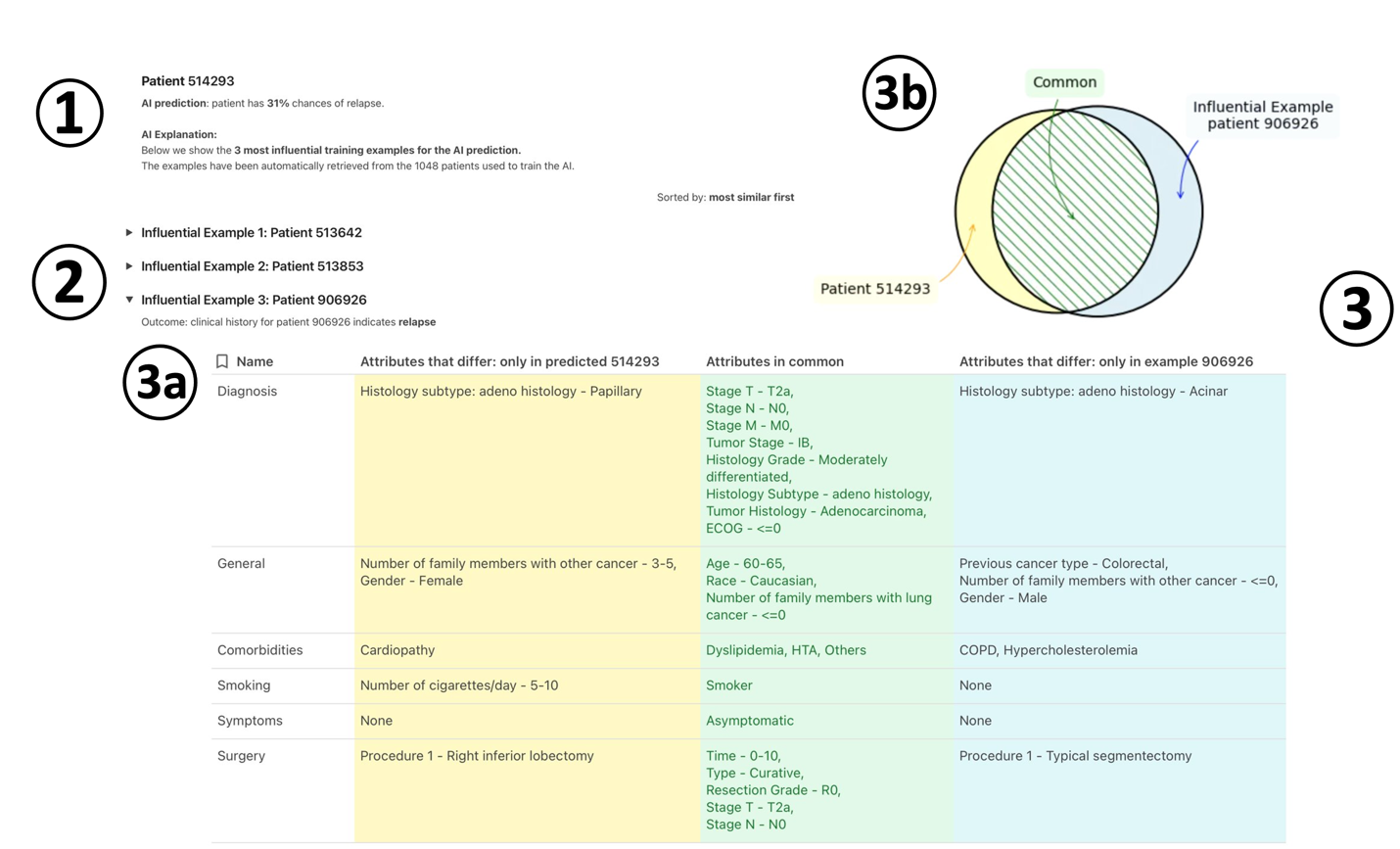}}

    \caption{Explanations provided by the pipeline for the two models for the same patient.}
    \label{fig:my_label}
\end{figure}

\section{DISCUSSION}

Regardless of the progress in cancer treatment in the past 10 years, 5-year survival remains at the level of 50$\%$ for resected NSCLC patients \cite{harpole_prognostic_1995}. Even in stage I, 20$\%$ of patients recurred within 5 year time \cite{harpole_prognostic_1995}.  

Predicting NSCLC relapse after surgery is essential for bespoke monitoring plans and adjuvant therapies \cite{wang_prediction_2017}. TNM-staging system is currently the only validated prognostic indicator for survival \cite{edge_american_2010}. Nevertheless, adoption of Machine Learning for other cancers substantially improves prediction accuracy \cite{xie_early_2021}. There exist no clinical, pathological, or molecular markers that predict risk of relapse with high accuracy other than the AJCC/International Staging System for lung cancer, which is i) focused on the nodal state affected and ii) highly dependent on the quality of the resection surgery and the number of resected nodes. Ideally, risk stratification among early-stage patients would be based on clinical and pathologic factors. This to administer adjuvant chemotherapy to stage I, high-risk-of-relapse patients and spare low-risk individuals from treatment toxicity \cite{wu_recurrence_2015} (except stage IB with > 4cm diameter).  Current standard practices offer adjuvant chemotherapy to patients with resected Stage IB to III NSCLC who have a good performance status (adjuvant chemotherapy improves overall survival in such patients \cite{goodgame_clinical_2008}).  A meta-analysis on 4,584 patients with completely resected NSCLC from five studies (ALPI, IALT, BLT, JBR10 and ANITA) shows that survival benefit of adjuvant treatment is limited to cisplatin-based chemotherapy in completely resected fit stage II-III patients (5$\%$ at 5 years) \cite{pignon_lung_2008}.

Machine learning's impact on improving treatment outcomes requires real-world validation. An analysis of 1,194 patients with NSCLC evaluates the prognostic signatures of quantitative imaging features, extracted with deep learning \cite{hosny_deep_2018}. In our work, we trained machine learning methods on 1,387 early-stage resected NSCLC patients’ features for  prediction with the best model reaching 76$\%$ accuracy.

We leverage factors associated with disease relapse, not on those associated only with overall survival. Because relapse analysis censors patients at the time of death without cancer relapse, the impact of non-cancer-related factors is minimized. Unlike survival analysis, we do not predict when relapse occurs, but whether it will happen or not. Our machine learning decision-support system does not rely on patients’ self-assessments. 

Our study is limited to a cohort of patients diagnosed and treated in Spain. Such demographic bias may affect applicability in hospitals with different practices. Our models are time-unaware: future work will address time support. It is worth mentioning we had no control on the data collection process and on input data inconsistencies.

Our previous work \cite{sameh_k_mohamed_et_al_predicting_2021} discussed baseline models that show clear promise for accurate lung cancer recurrence predictions, that are based not on population statistics but on individual features of the patients. We now add graph machine learning and explainable artificial intelligence methods to make sure clinicians make informed decisions when adopting the system in patient follow-ups \cite{sung_global_2021, noauthor_cancer_nodate}.

Our work would benefit from patient-level molecular data and medical imaging tumour assessment results. Mounting evidence shows that CT image features have high diagnostic and predictive values in clinical pathologic staging of diseases and clinical outcomes \cite{yu_development_2018, thawani_radiomics_2018, yu_development_2018-1}. Additionally, previous studies have described the development of gene-expression, protein, and messenger RNA profiles that are associated in some cases with the outcome of lung cancer \cite{aramburu_combined_2015, berrar_survival_2005}.

\section{CONCLUSIONS}

We use tabular and graph machine learning for objective and reproducible early-stage NSCLC recurrence. This enables new lung cancer stratification methods based on a personalised relapse risk score.

With further prospective and multi-site validation and the addition of radiological and genomic features to the patient´s profile, this prognostic model can serve as a predictive decision support tool for deciding the use of adjuvant treatments in early-stage lung cancer.

\bibliographystyle{IEEEtran}
\bibliography{bibliography}

\end{document}


\pagenumbering{Roman}
\title{Supplementary Material}
\section*{Supplementary Material}
\begin{enumerate}[]
    \item A Exploratory Data Analysis
\item B Glossary
\item C Additional ML Experiments
\item E Class-Based Evaluation
\item F Confusion Matrices for Models
\item G Receiver Operating Characteristic (ROC) Curves for Models
\item H Waterfall Plots for Tabular Models
\item I Metrics
\item J Model Parameters
\item H Time-based Relapse Prediction*
\end{enumerate}

\newpage
\section*{A Exploratory Data Analysis}

\textit{Missing Values}

\begin{table}[H]
\begin{tabularx}{\textwidth}{|
p{\dimexpr 0.254\linewidth-2\tabcolsep-2\arrayrulewidth}|
p{\dimexpr 0.329\linewidth-2\tabcolsep-\arrayrulewidth}|
p{\dimexpr 0.184\linewidth-2\tabcolsep-\arrayrulewidth}|
p{\dimexpr 0.232\linewidth-2\tabcolsep-\arrayrulewidth}|} \hline 
\textbf{Category} & \textbf{Attribute} & \textbf{Number of Missing Values} & \textbf{Percentage} \\\hline 
\multirow[t]{2}{=}{Smoking}  & Number of cigarettes packs/year & 400 & 28.84\% \\\cline{2-4}
 & Number of cigarettes/day & 444 & 32.01\% \\\hline 
\multirow[t]{2}{=}{General}  & Age & 4 & 0.29\% \\\cline{2-4}
 & Previous cancer type & 1084 & 78.15\% \\\hline 
\multirow[t]{8}{=}{Diagnosis}  & Histology Grade & 578 & 41.67\% \\\cline{2-4}
 & Histology Subtype & 790 & 56.96\% \\\cline{2-4}
 & ECOG & 3 & 0.22\% \\\cline{2-4}
 & ALK IHQ & 1070 & 77.14\% \\\cline{2-4}
 & PD-L1 & 1099 & 79.24\% \\\cline{2-4}
 & Histology subtype: adeno histology & 577 & 41.6\% \\\cline{2-4}
 & Biomarker type & 1348 & 97.19\% \\\cline{2-4}
 & EGFR negative & 995 & 71.74\% \\\hline 
\multirow[t]{3}{=}{Chemotherapy}  & Type & 690 & 49.75\% \\\cline{2-4}
 & Start Time {[}months{]} & 729 & 52.56\% \\\cline{2-4}
 & Regimen & 692 & 49.89\% \\\hline 
\multirow[t]{7}{=}{Surgery}  & Procedure 1 & 85 & 6.13\% \\\cline{2-4}
 & Perioperative Death & 80 & 5.77\% \\\cline{2-4}
 & Time & 103 & 7.43\% \\\cline{2-4}
 & Type & 107 & 7.71\% \\\cline{2-4}
 & Resection Grade & 174 & 12.55\% \\\cline{2-4}
 & Stage T & 117 & 8.44\% \\\cline{2-4}
 & Stage N & 117 & 8.44\% \\\hline 
\multirow[t]{5}{=}{Radiotherapy}  & Duration {[}days{]} & 1332 & 96.03\% \\\cline{2-4}
 & Type & 1324 & 95.46\% \\\cline{2-4}
 & Area & 1323 & 95.39\% \\\cline{2-4}
 & Dose & 1333 & 96.11\% \\\cline{2-4}
 & Fractioning & 1338 & 96.47\% \\\hline 
\end{tabularx}
\end{table}
\begin{center}
    \textit{Table 4: List of missing values according to group of features.}
\end{center}

\subsection*{Radiotherapy}
\begin{figure}[H]
\includegraphics[width=1\textwidth]{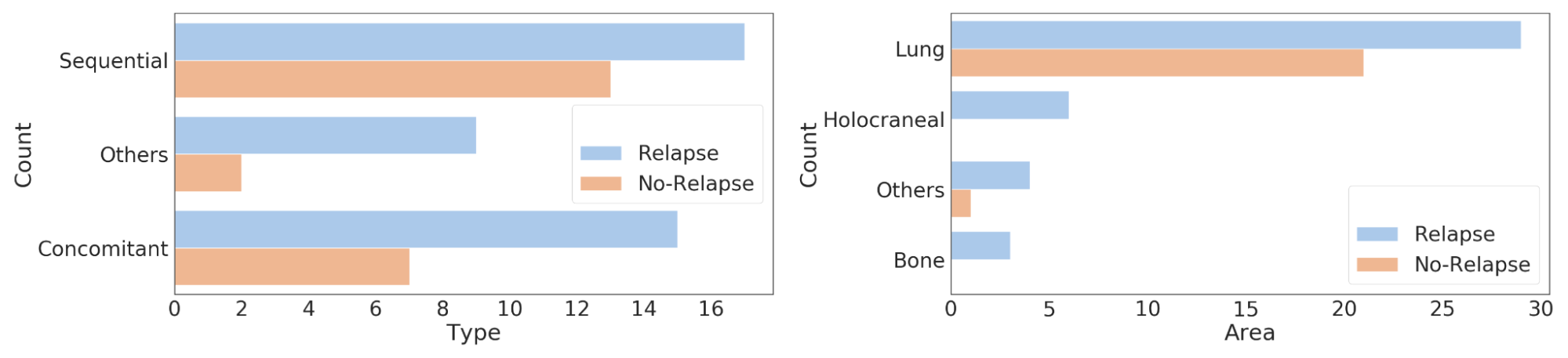}\caption{\textit{Figure 4}}
\label{fig:1}
\end{figure}
\subsection*{Chemotherapy}
\begin{figure}[H]
\includegraphics[width=1\textwidth]{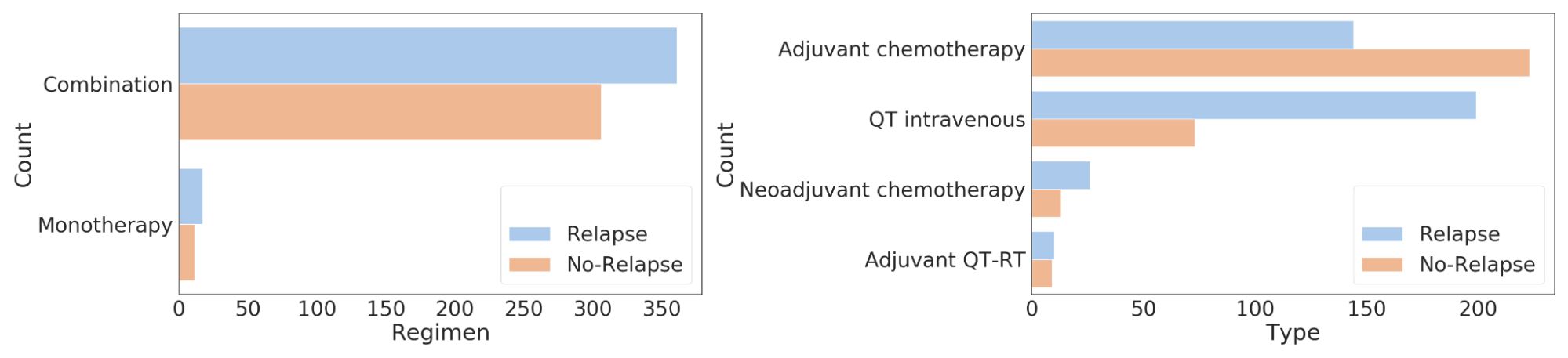}\caption{\textit{Figure 5}}
\label{fig:2}
\end{figure}

\subsection*{Surgery}
\begin{figure}[H]
\includegraphics[width=1\textwidth]{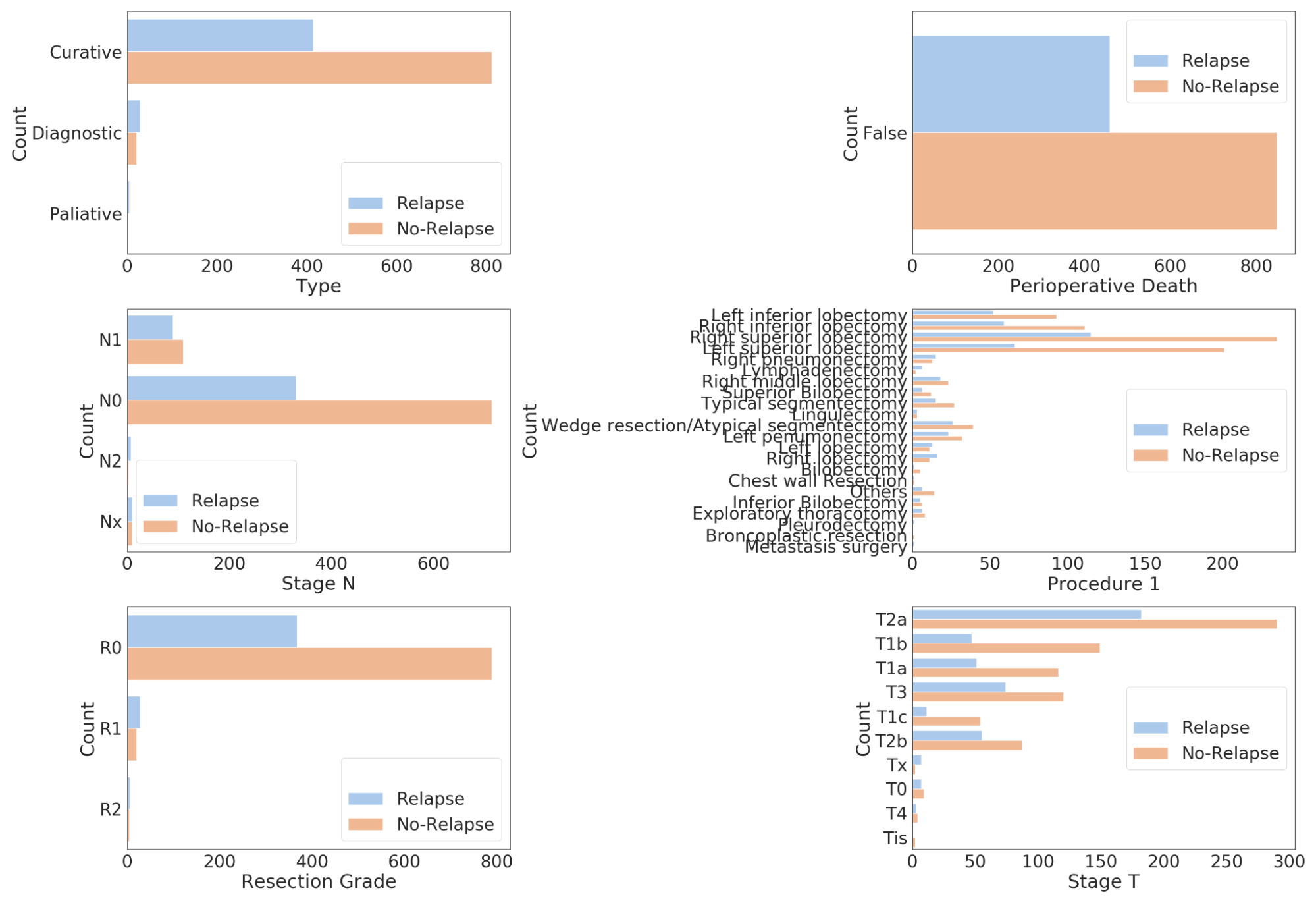}\caption{\textit{Figure 6}}
\label{fig:3}
\end{figure}
\subsection*{Diagnosis}
\begin{figure}[H]
    \centering
\includegraphics[width=1\textwidth]{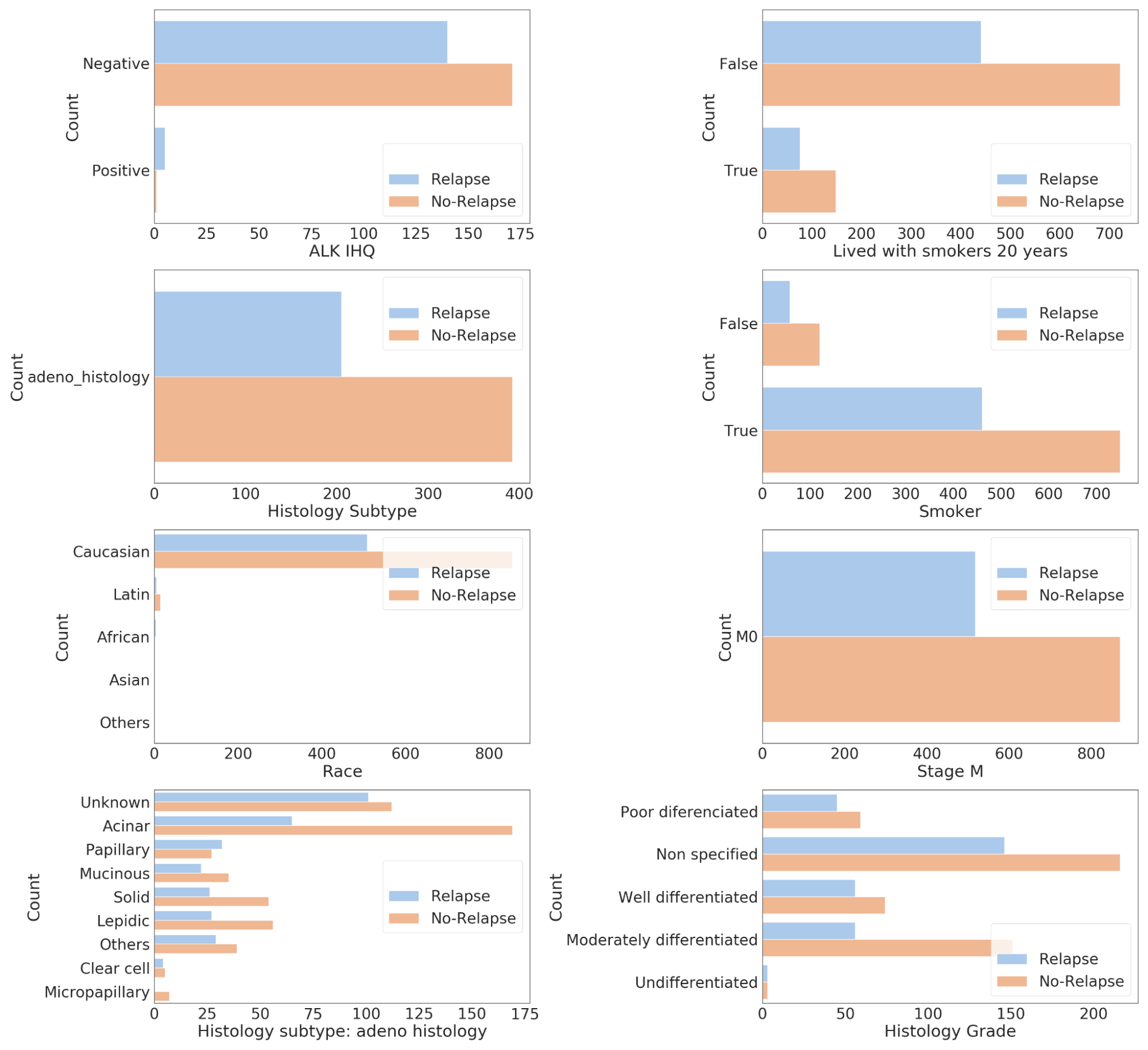}
    \caption{\textit{Figure 7}}
    \label{fig:my_label}
\end{figure}

\begin{figure}[H]
\includegraphics[width=1\textwidth]{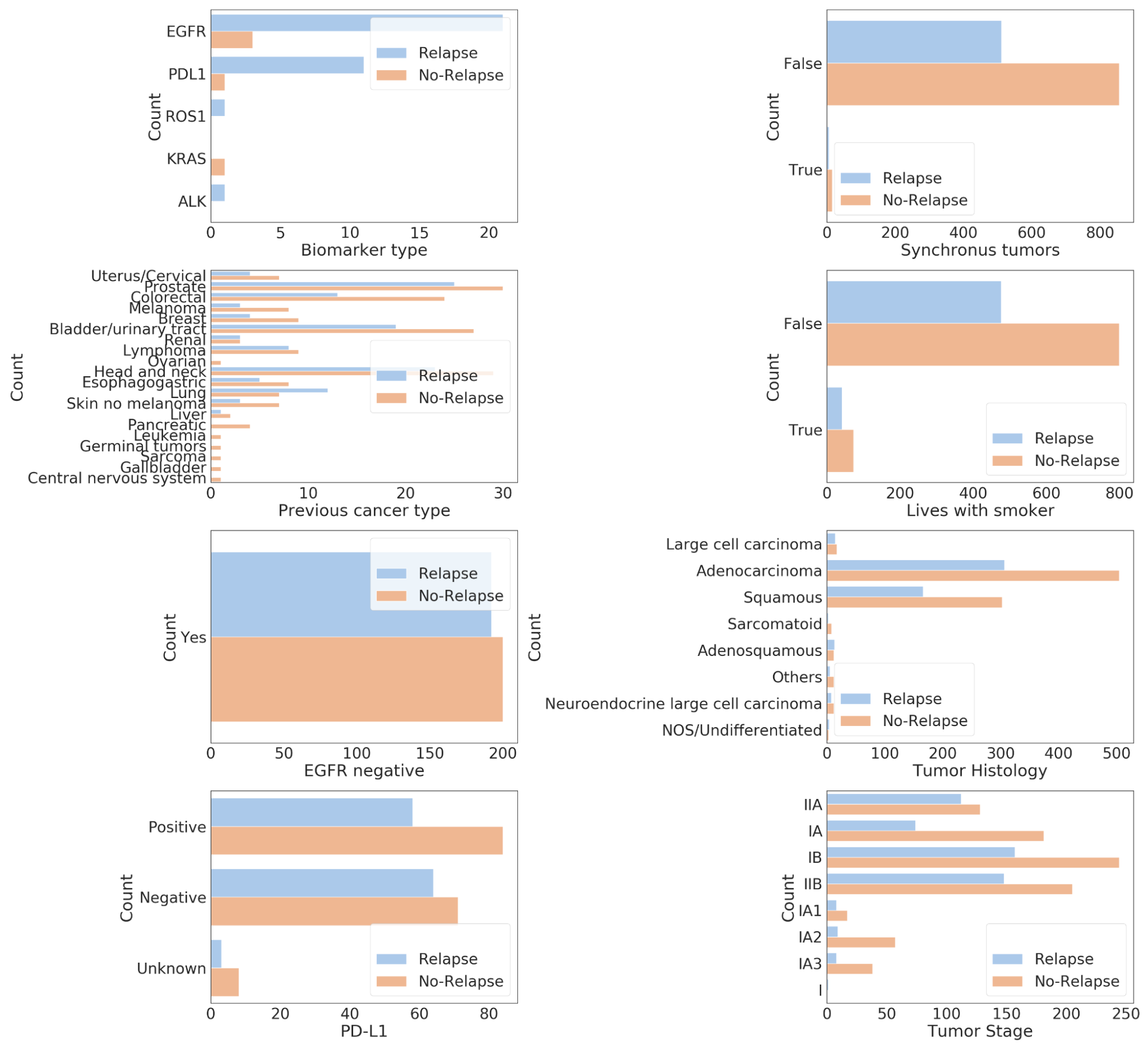}\caption{\textit{Figure 8}}
\label{fig:4}
\end{figure}
\subsection*{Comorbidities}
\begin{figure}[H]
    \centering
\includegraphics[width=1\textwidth]{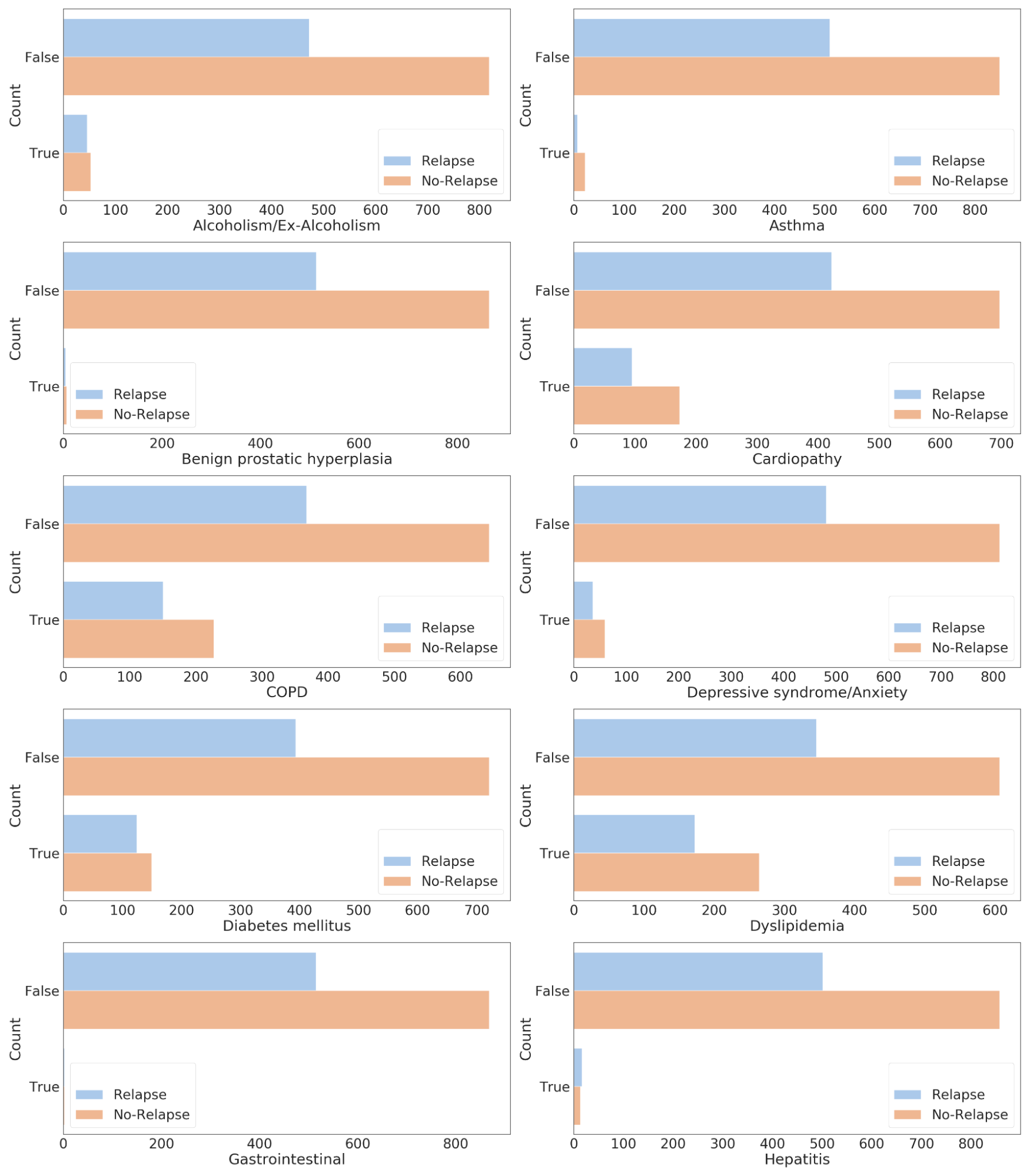}
    \caption{\textit{Figure 9}}
    \label{fig:my_label}
\end{figure}

\begin{figure}[H]
\includegraphics[width=1\textwidth]{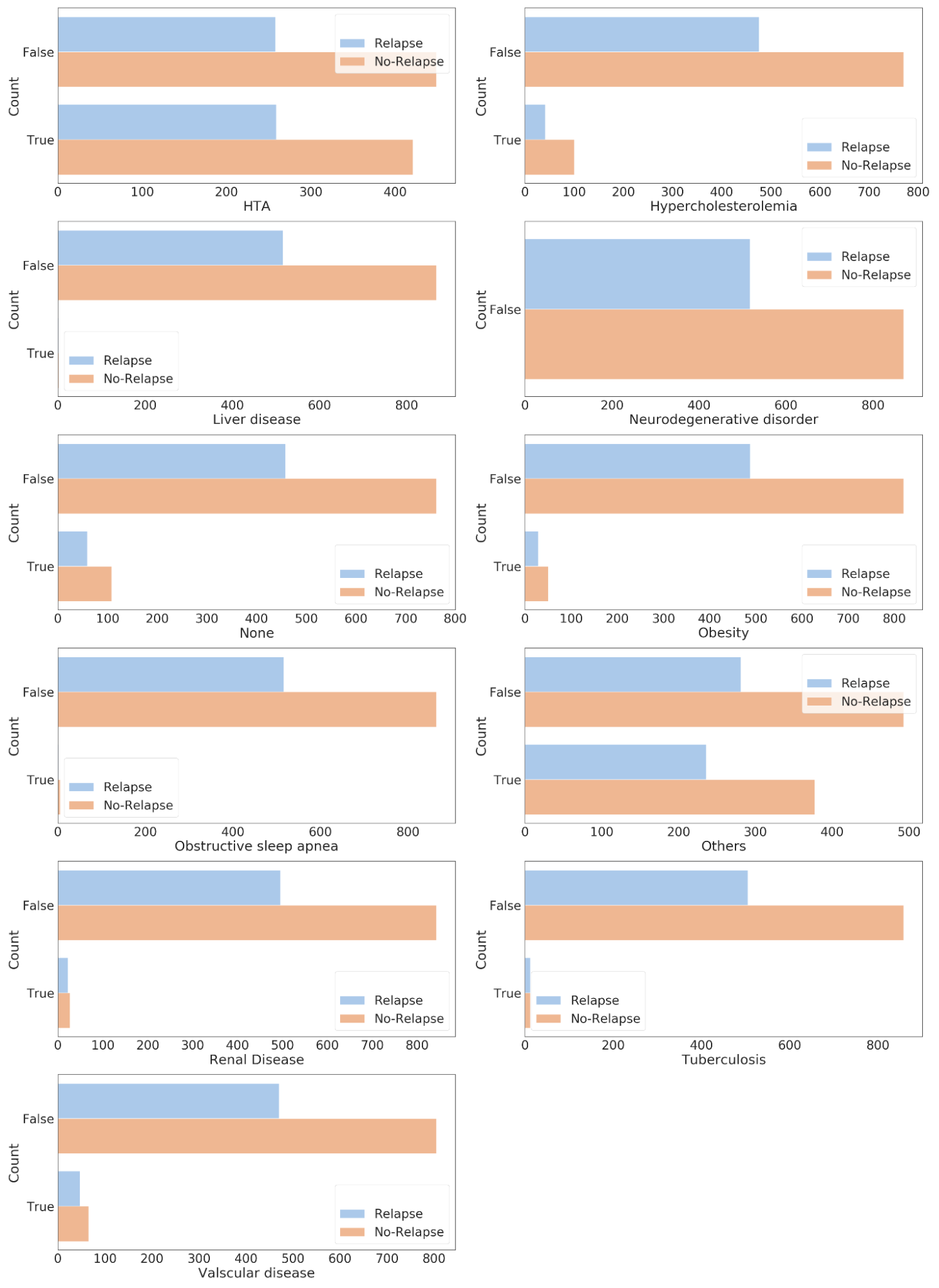}\caption{\textit{Figure 10}}
\label{fig:5}
\end{figure}
\subsection*{Symptoms}
\begin{figure}[H]
\includegraphics[width=1\textwidth]{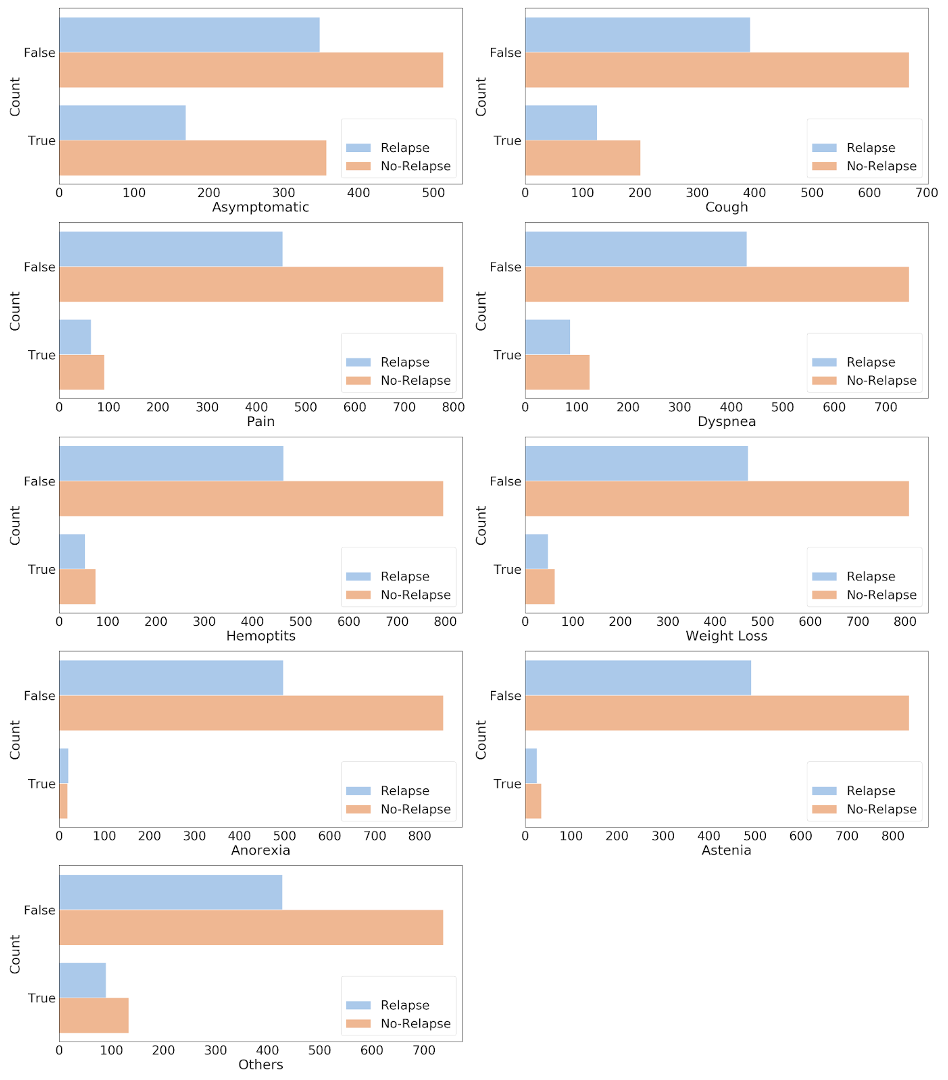}\caption{\textit{Figure 11}}
\label{fig:6}
\end{figure}

\begin{figure}
    \centering
    \includegraphics[width=1\textwidth]{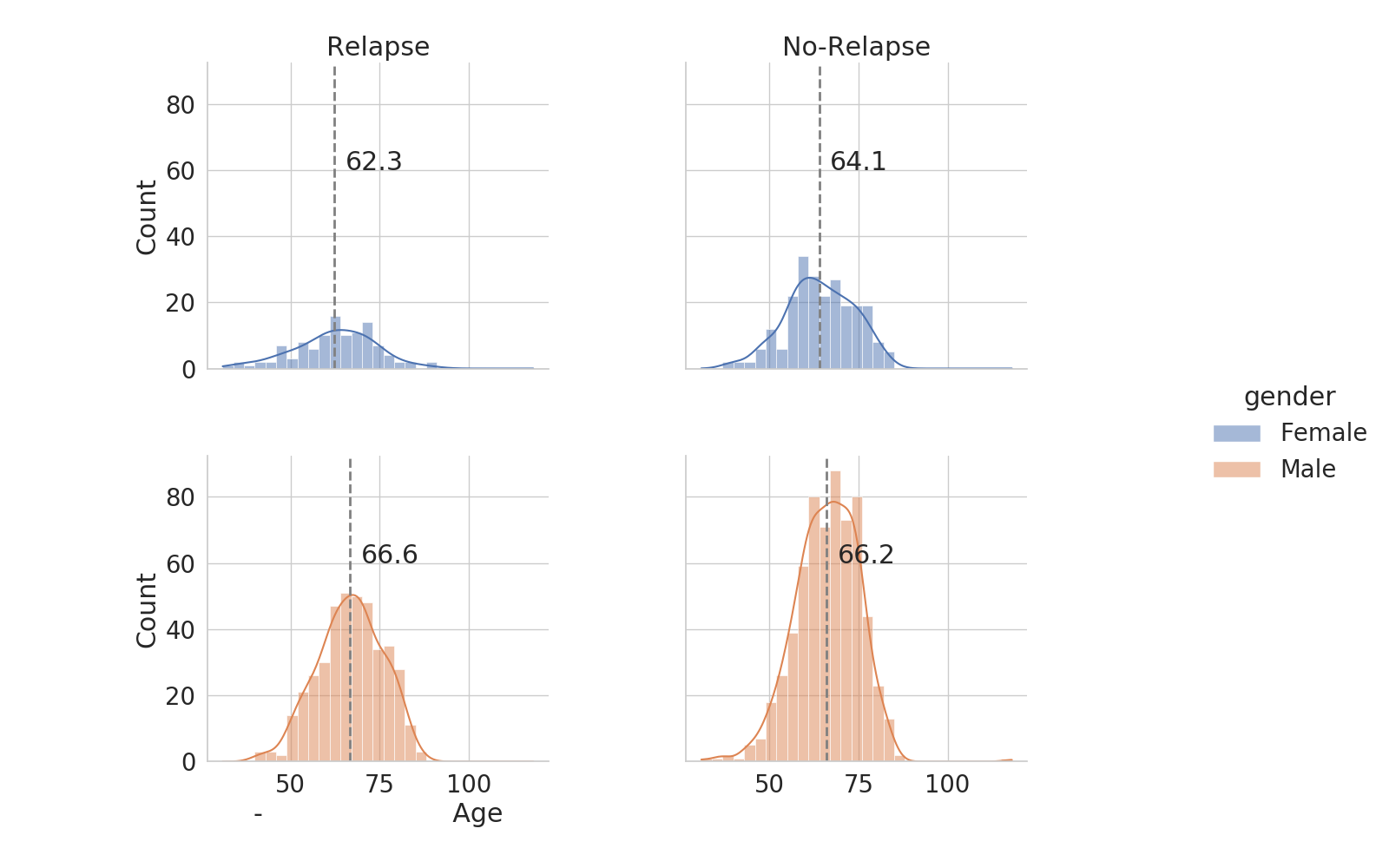}
    \caption{\textit{Figure 12: Histograms of patient age according to their gender and whether they had a relapse (Recurrence) or not (Survival). The first row represents female patients (blue) and the second male patients (orange). For every histogram, the mean value is annotated and marked with a dashed line. Plots are overlaid with Kernel Density Estimate (KDE) plots for clarity.}}
    \label{fig:my_label}
\end{figure}

\begin{figure}
     \centering
     \begin{subfigure}[b]{0.47\textwidth}
         \centering
        \includegraphics[width=1\textwidth]{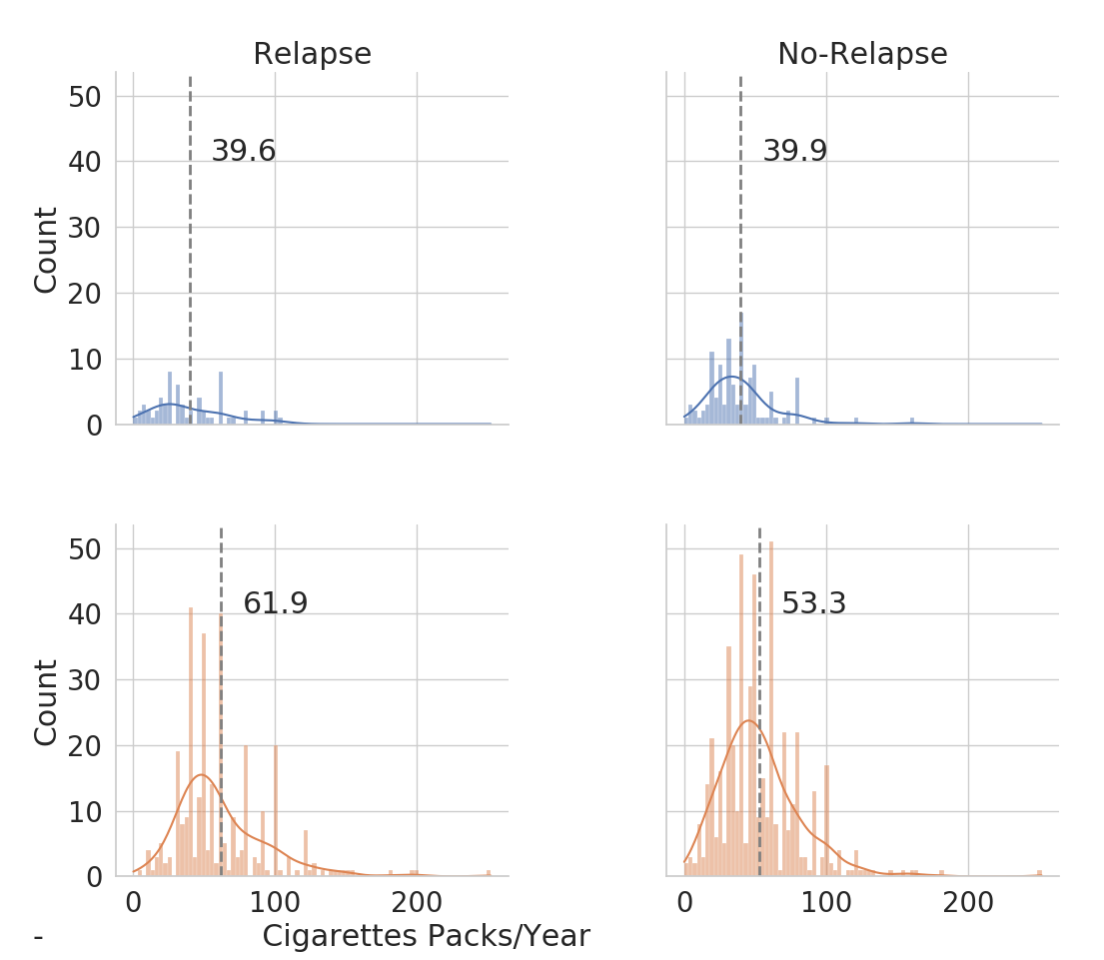}

     \end{subfigure}
     \hfill
     \begin{subfigure}[b]{0.47\textwidth}
         \centering
         \includegraphics[width=1\textwidth]{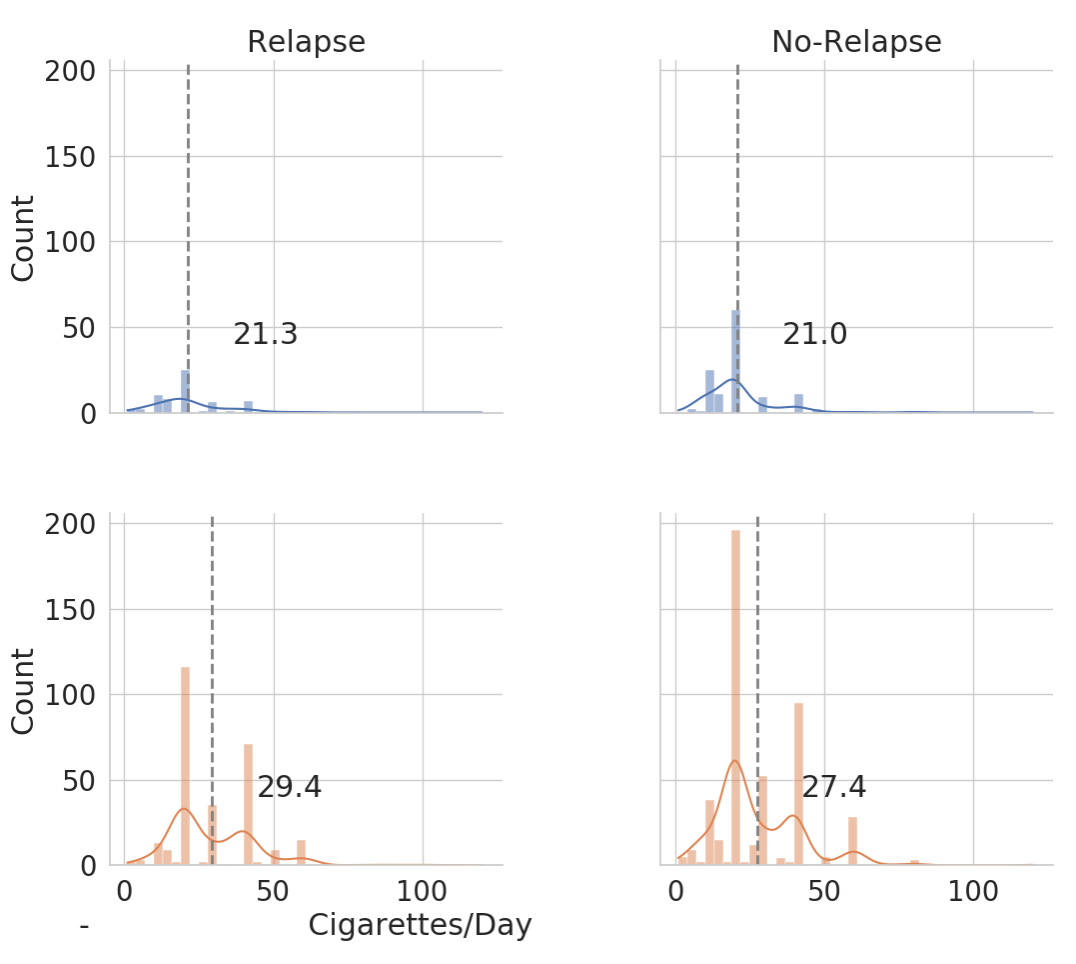}

     \end{subfigure}
\end{figure}

\begin{figure}
     \centering
     \begin{subfigure}[b]{0.47\textwidth}
         \centering
        \includegraphics[width=1\textwidth]{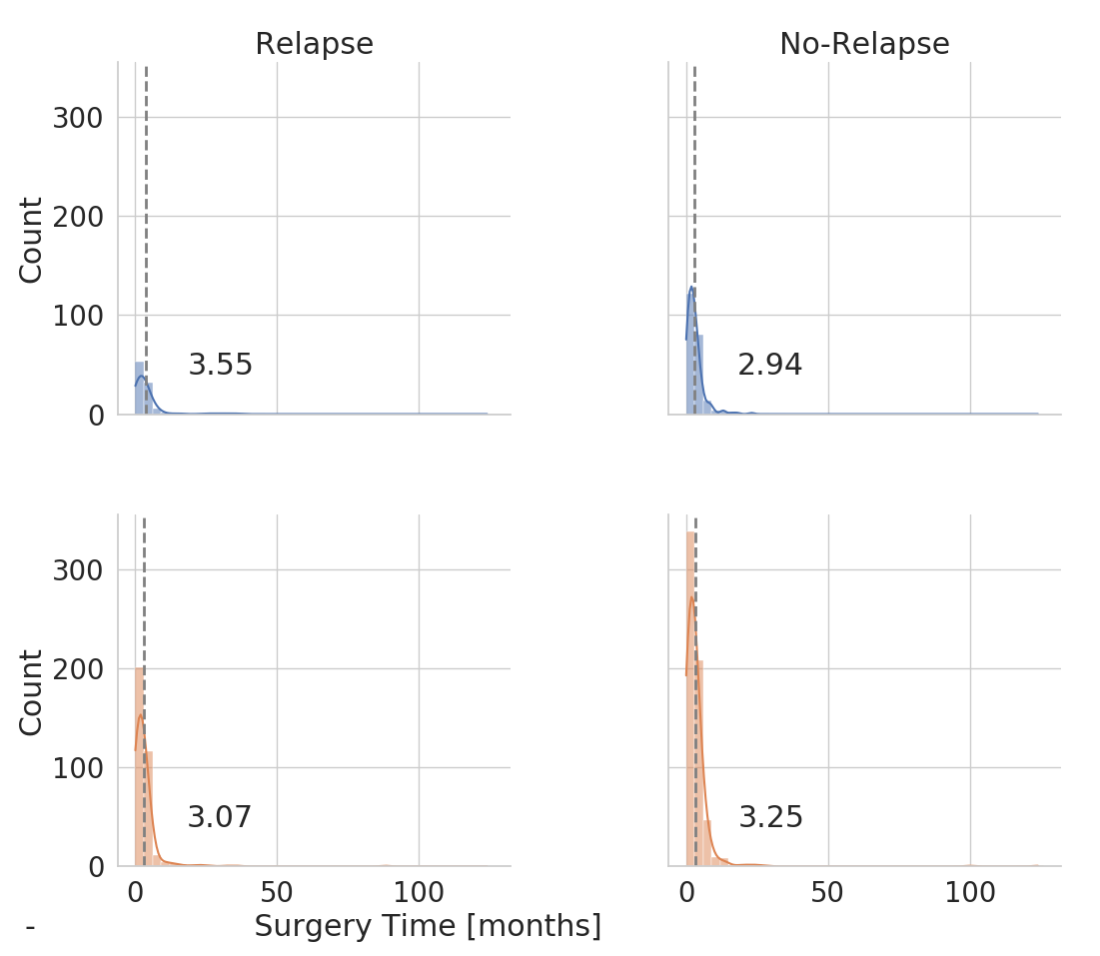}

     \end{subfigure}
     \hfill
     \begin{subfigure}[b]{0.47\textwidth}
         \centering
         \includegraphics[width=1\textwidth]{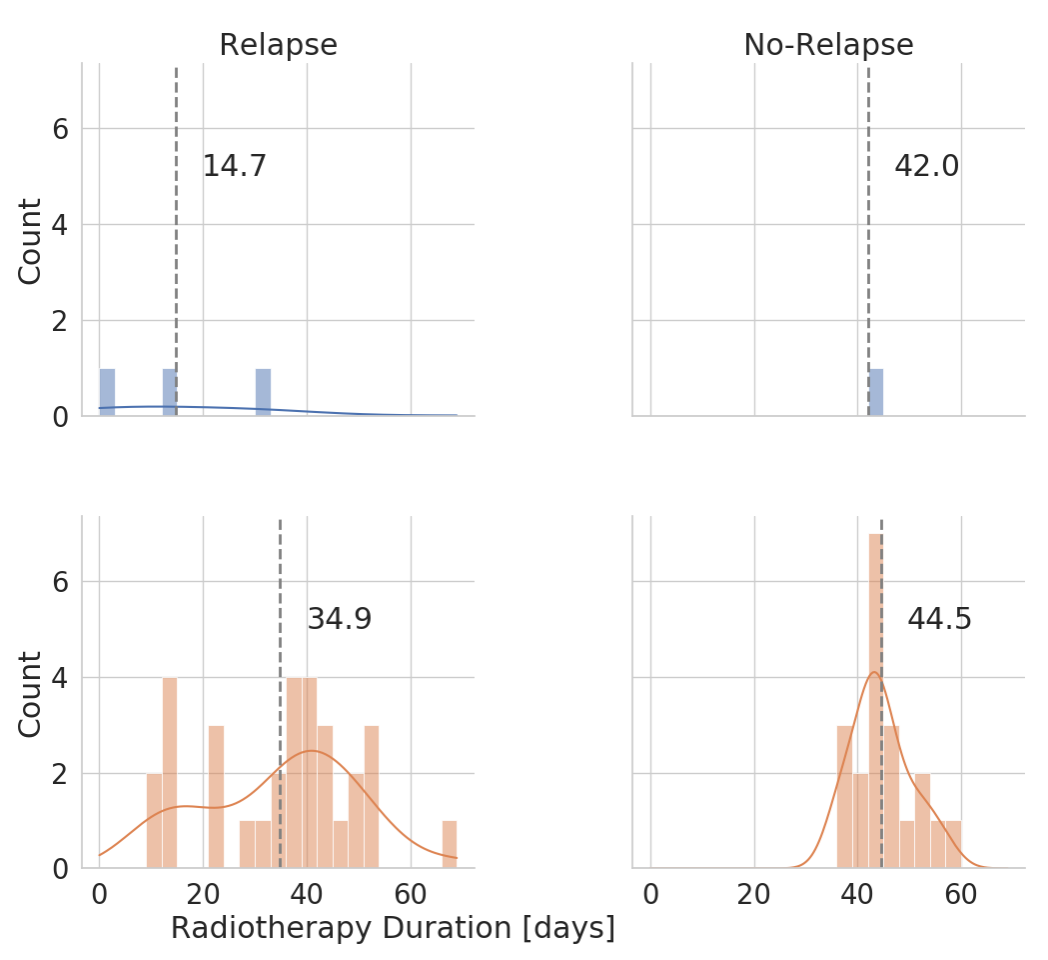}

     \end{subfigure}
\end{figure}
\includegraphics[width=1\textwidth]{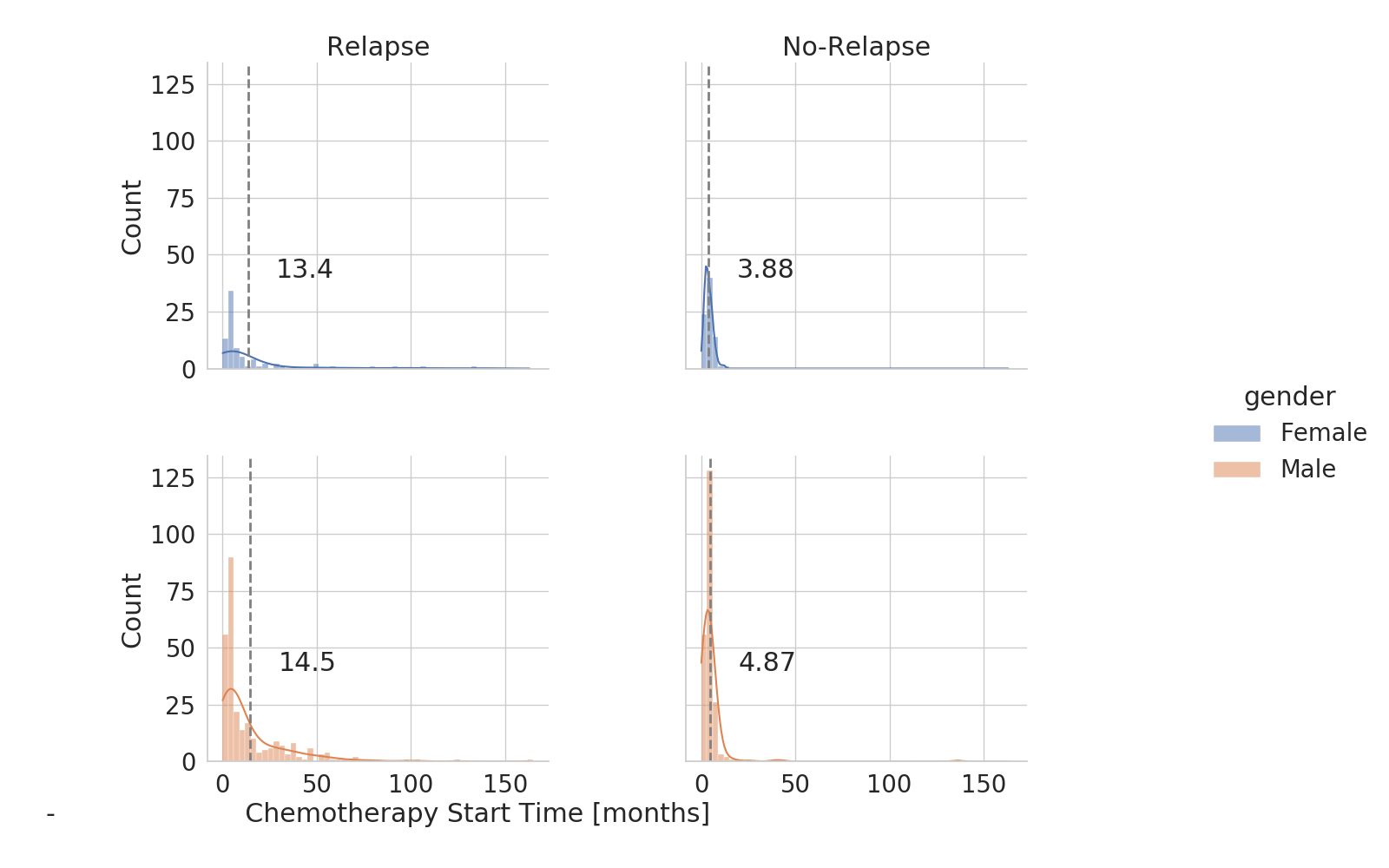}

\textit{Figure 13: Histograms of patient numerical features (cigarettes packs/year, cigarettes/day, surgery time, chemotherapy start time, and radiotherapy duration) according to their gender and whether they had a relapse (Recurrence) or not (Survival). The first row represents female patients (blue) and the second male patients (orange). For every histogram, the mean value is annotated and marked with a dashed line. Plots are overlaid with Kernel Density Estimate (KDE) plots for clarity.}

\begin{figure}[H]
     \centering
     \begin{subfigure}[b]{0.3\textwidth}
         \centering
        \includegraphics[width=1\textwidth]{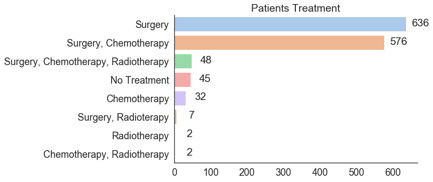}
        \caption{ Patients according to treatments.}

     \end{subfigure}
     \hfill
     \begin{subfigure}[b]{0.3\textwidth}
         \centering
         \includegraphics[width=1\textwidth]{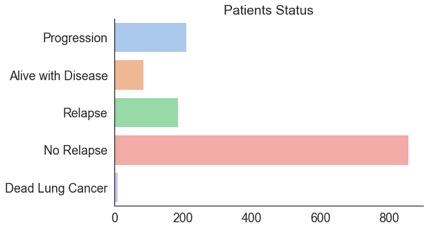}
\caption{Source of ground truth labels for patients.}
     \end{subfigure}
     \hfill
     \begin{subfigure}[b]{0.3\textwidth}
         \centering
         \includegraphics[width=1\textwidth]{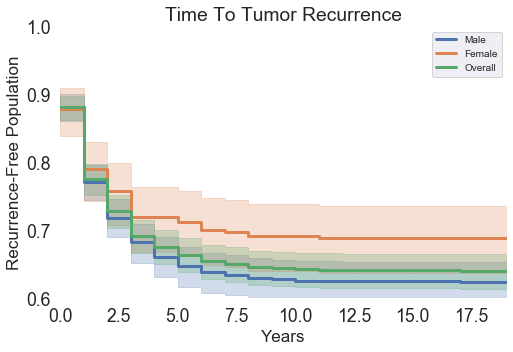}
\caption{Kaplan Meier curve for Time to Tumor Recurrence in males, females and overall.}
     \end{subfigure}
     \caption{\centering{}\textit{Figure 14. a) Patients treatment b) Labels c) Time to Tumor Recurrence.}}
\end{figure}

\section*{B Glossary}

\textit{Concise explanations for ML concepts included in the paper in alphabetical order.}

\textbf{Baseline models:} In general, any predictive model that is used as a starting point for developing production-ready solutions to the given problem. Can be a random baseline (i.e. predictions that could be made by tossing a coin) or more sophisticated algorithms that do produce better-than-random results already and can be already used in theory, but are by no means the expected end solution.

\textbf{Cox proportional hazards model:} A statistical method widely used for risk-based patient stratification by 

predicting the likelihood that an adversity occurs in patients within a given timeframe.

\textbf{Explainable AI} (or also interpretable \textbf{machine learning}): Predictive algorithms (largely based on machine learning in the context of this deliverable) that can produce not only predictions but also their explanations. Essential in healthcare where decisions always have to be accountable and anchored in a broader biomedical context.

\textbf{Explainable models:} Predictive models that can also explain their predictions.

\textbf{Knowledge graph:} A machine-readable representation of knowledge bases in a graph-based form, i.e., using relationships between entities relevant to a specific domain and/or problem. It can be efficiently used for knowledge base completion by means of predicting new links in the graph, and thus support making predictions.

\textbf{Learning from tabular data:} Machine learning performed on data in the form of a matrix, where the rows can be imagined to correspond to training examples while the columns represent specific values of features characteristic for the examples.

\textbf{Machine learning:} A computer science discipline that, in a nutshell, aims at developing predictive algorithms that can automatically improve their performance. They learn to make predictions about unseen domain data by being trained on data approximating a sample of the domain knowledge.

\textbf{Predictive models:} Machine learning algorithms that have been trained and can make predictions in the given domain.

\textbf{Relational learning:} Machine learning performed on data in the form of a (knowledge) graph that makes use of not only the data features (i.e. links between entities) but also of the inherent network structure of the data set.

\section*{C Additional ML Experiments}

\begin{table}[H]
\begin{tabularx}{\textwidth}{|
p{\dimexpr 0.088\linewidth-2\tabcolsep-2\arrayrulewidth}|
p{\dimexpr 0.194\linewidth-2\tabcolsep-\arrayrulewidth}|
p{\dimexpr 0.119\linewidth-2\tabcolsep-\arrayrulewidth}|
p{\dimexpr 0.119\linewidth-2\tabcolsep-\arrayrulewidth}|
p{\dimexpr 0.119\linewidth-2\tabcolsep-\arrayrulewidth}|
p{\dimexpr 0.119\linewidth-2\tabcolsep-\arrayrulewidth}|
p{\dimexpr 0.125\linewidth-2\tabcolsep-\arrayrulewidth}|
p{\dimexpr 0.119\linewidth-2\tabcolsep-\arrayrulewidth}|} \hline 
\centering\arraybackslash{} & \centering\arraybackslash{}\textbf{Model} & \centering\arraybackslash{}\textbf{Accuracy} & \centering\arraybackslash{}\textbf{P} & \centering\arraybackslash{}\textbf{R} & \centering\arraybackslash{}\textbf{F1} & \centering\arraybackslash{}\textbf{Average Precision} & \centering\arraybackslash{}\textbf{AUC-ROC} \\\hline 
\centering\arraybackslash{} & \multicolumn{7}{p{\dimexpr 0.914\linewidth-2\tabcolsep-\arrayrulewidth}|}{\centering\arraybackslash{}\centering\arraybackslash{}\textbf{Relapse prediction with diagnosis features}} \\\hline 
\centering\arraybackslash{}1. & \centering\arraybackslash{}ComplEx-N3 & \centering\arraybackslash{}0.44 & \centering\arraybackslash{} 0.44 & \centering\arraybackslash{} 0.41 & \centering\arraybackslash{}0.42 & \centering\arraybackslash{} 0.47 & \centering\arraybackslash{} 0.44 \\\hline 
\centering\arraybackslash{} & \multicolumn{7}{p{\dimexpr 0.914\linewidth-2\tabcolsep-\arrayrulewidth}|}{\centering\arraybackslash{}\centering\arraybackslash{}\textbf{Relapse prediction after surgery}} \\\hline 
\centering\arraybackslash{}2. & \centering\arraybackslash{}ComplEx-N3 & \centering\arraybackslash{}0.68 & \centering\arraybackslash{} 0.66 & \centering\arraybackslash{} \textbf{0.74}  & \centering\arraybackslash{}\textbf{0.7} & \centering\arraybackslash{} 0.62 & \centering\arraybackslash{}0.68 \\\hline 
\centering\arraybackslash{} & \multicolumn{7}{p{\dimexpr 0.914\linewidth-2\tabcolsep-\arrayrulewidth}|}{\centering\arraybackslash{}\centering\arraybackslash{}\textbf{Relapse prediction after surgery and treatment}} \\\hline 
\centering\arraybackslash{}3. & \centering\arraybackslash{}ComplEx-N3 & \centering\arraybackslash{}\textbf{0.69} & \centering\arraybackslash{}\textbf{0.72} & \centering\arraybackslash{}0.61 & \centering\arraybackslash{}0.66 & \centering\arraybackslash{}\textbf{0.63} & \centering\arraybackslash{}\textbf{0.69} \\\hline 
\end{tabularx}
\end{table}

\begin{figure}[H]
\includegraphics[width=1\textwidth]{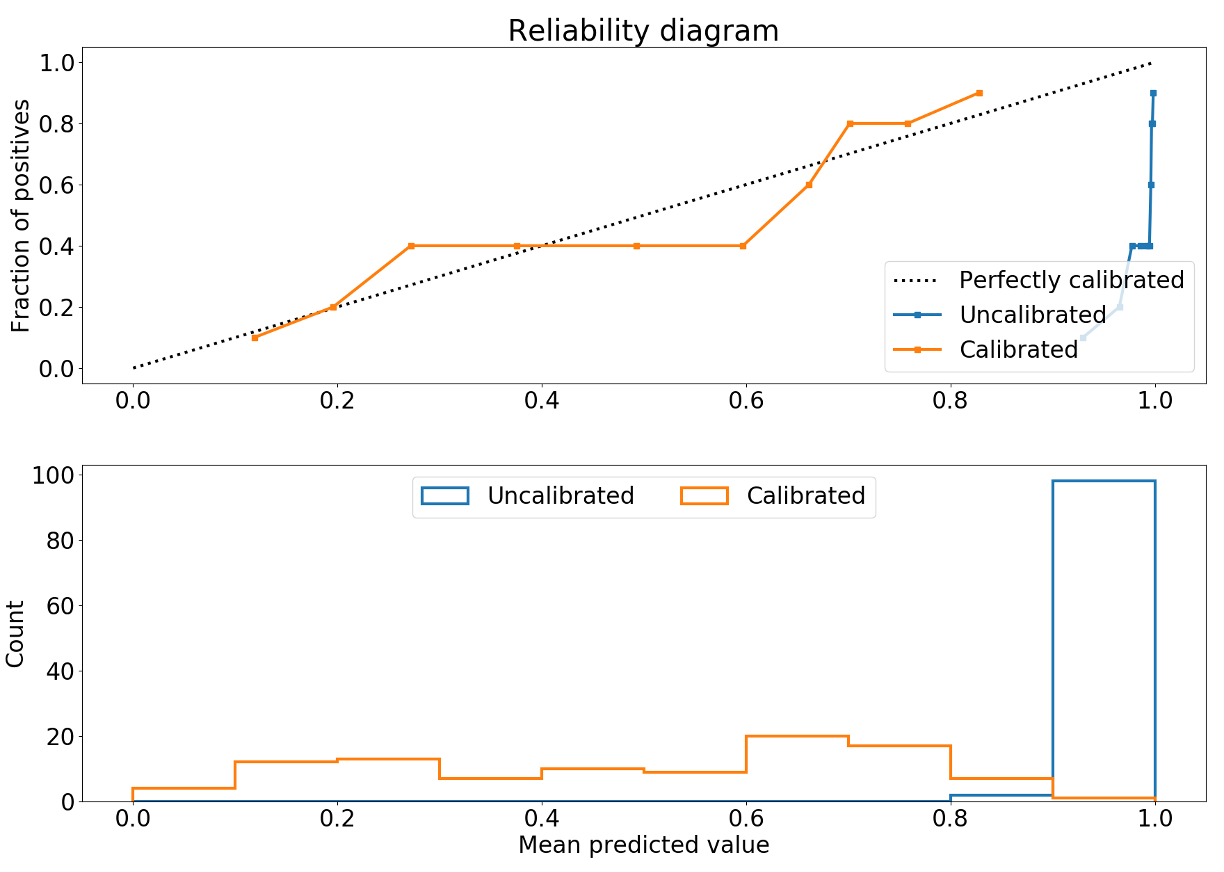}\caption{\textit{Figure 15: Reliability diagram presents how calibrated scores differ from the raw scores returned as a prediction by the model. Calibrated scores are presented in orange, uncalibrated scores in blue. Dashed line serves as a reference for the case when the scores would have been perfectly calibrated.}}
\label{fig:7}
\end{figure}
\newpage
\section*{D Features List}
\begin{table}[H]
\caption{\centering{}\textbf{Full features list generated by the pipeline, including training features, identification, label, and features with missing values filtered out before training.}}
\begin{tabularx}{\textwidth}{|
p{\dimexpr 0.403\linewidth-2\tabcolsep-2\arrayrulewidth}|
p{\dimexpr 0.223\linewidth-2\tabcolsep-\arrayrulewidth}|
p{\dimexpr 0.243\linewidth-2\tabcolsep-\arrayrulewidth}|
p{\dimexpr 0.131\linewidth-2\tabcolsep-\arrayrulewidth}|} \hline 
\textbf{Features} & \textbf{Name} & \textbf{Description} & \centering\arraybackslash{}\textbf{Filter} \\\hline 
\multicolumn{4}{|p{\dimexpr 1\linewidth-2\tabcolsep-2\arrayrulewidth}|}{\centering\arraybackslash{}\centering\arraybackslash{}\textbf{Features filtered with the selection criteria} \newline (only values specified in the Filter column were included in the model)} \\\hline 
diagnosis\_stage\_t & Stage T & T stage of the patient's tumor. Relates to the primary tumor's size. & T1a, T1b, T1c, T2a, T2b, T0, Tx, Tis, T3 \\\hline 
diagnosis\_stage\_n & Stage N & N stage of the patient's tumor. Relates to the degree of spread to lymph nodes. & N0, N1 \\\hline 
diagnosis\_stage\_m & Stage M & M stage of the patient's tumor. Relates to the presence of metastasis. & M0 \\\hline 
chemotherapy@t1\_type & Type & Chemotherapy type conducted. & Adjuvant chemotherapy,  \newline CHT   intravenous,  \newline Adjuvant \newline   CHT-RT  \newline Neoadjuvant\\\hline 
tumor\_stage & Tumor Stage & Stage associated to the patient (I, II, III, IV,...). & I, IA, IA1, IA2, IA3, IB, II, IIA, IIB \\\hline 

\multicolumn{4}{|p{\dimexpr 1\linewidth-2\tabcolsep-2\arrayrulewidth}|}{\centering\arraybackslash{}\centering\arraybackslash{}\textbf{Comorbidities}} \\\hline 
comorbidity\_alcoholism\_or\_ex\_alcoholism & Alcoholism/Ex-Alcoholism & \multirow[t]{4}{=}{Name of the comorbidity associated to the patient.}  & \centering\arraybackslash{}- \\\cline{1-2}\cline{4-4}
comorbidity\_asthma & Asthma &  & \centering\arraybackslash{}- \\\cline{1-2}\cline{4-4}
comorbidity\_benign\_prostatic\_hyperplasia & Benign prostatic hyperplasia &  & \centering\arraybackslash{}- \\\cline{1-2}\cline{4-4}
comorbidity\_cardiopathy & Cardiopathy &  & \centering\arraybackslash{}- \\\hline 
comorbidity\_copd & COPD & Chronic obstructive pulmonary disease & \centering\arraybackslash{}- \\\hline 
comorbidity\_depressive\_syndrome\_or\_anxiety & Depressive syndrome/Anxiety & \multirow[t]{9}{=}{Name of the comorbidity associated to the patient.}  & \centering\arraybackslash{}- \\\cline{1-2}\cline{4-4}
comorbidity\_diabetes\_mellitus & Diabetes mellitus &  & \centering\arraybackslash{}- \\\cline{1-2}\cline{4-4}
comorbidity\_dyslipidemia & Dyslipidemia &  & \centering\arraybackslash{}- \\\cline{1-2}\cline{4-4}
comorbidity\_gastrointestinal & Gastrointestinal &  & \centering\arraybackslash{}- \\\cline{1-2}\cline{4-4}
comorbidity\_hepatitis & Hepatitis &  & \centering\arraybackslash{}- \\\cline{1-2}\cline{4-4}
comorbidity\_hta & HTA &  & \centering\arraybackslash{}- \\\cline{1-2}\cline{4-4}
comorbidity\_hypercholesterolemia & Hypercholesterolemia &  & \centering\arraybackslash{}- \\\cline{1-2}\cline{4-4}
comorbidity\_liver\_disease & Liver disease &  & \centering\arraybackslash{}- \\\cline{1-2}\cline{4-4}
comorbidity\_neurodegenerative\_disorder & Neurodegenerative disorder &  & \centering\arraybackslash{}- \\\hline 
comorbidity\_none & None & \multirow[t]{7}{=}{Name of the comorbidity associated to the patient.}  & \centering\arraybackslash{}- \\\cline{1-2}\cline{4-4}
comorbidity\_obesity & Obesity &  & \centering\arraybackslash{}- \\\cline{1-2}\cline{4-4}
comorbidity\_obstructive\_sleep\_apnea & Obstructive sleep apnea &  & \centering\arraybackslash{}- \\\cline{1-2}\cline{4-4}
comorbidity\_others & Others &  & \centering\arraybackslash{}- \\\cline{1-2}\cline{4-4}
comorbidity\_renal\_disease & Renal Disease &  & \centering\arraybackslash{}- \\\cline{1-2}\cline{4-4}
comorbidity\_tuberculosis & Tuberculosis &  & \centering\arraybackslash{}- \\\cline{1-2}\cline{4-4}
comorbidity\_valscular\_disease & Valscular disease &  & \centering\arraybackslash{}- \\\hline 
\end{tabularx}

\end{table}

\begin{table}[H]
\begin{tabularx}{\textwidth}{|
p{\dimexpr 0.403\linewidth-2\tabcolsep-2\arrayrulewidth}|
p{\dimexpr 0.223\linewidth-2\tabcolsep-\arrayrulewidth}|
p{\dimexpr 0.243\linewidth-2\tabcolsep-\arrayrulewidth}|
p{\dimexpr 0.131\linewidth-2\tabcolsep-\arrayrulewidth}|} \hline 
\textbf{Features} & \textbf{Name} & \textbf{Description} & \centering\arraybackslash{}\textbf{Filter} \\\hline 
\multicolumn{4}{|p{\dimexpr 1\linewidth-2\tabcolsep-2\arrayrulewidth}|}{\centering\arraybackslash{}\centering\arraybackslash{}\textbf{Smoking Features}} \\\hline 
smoker & Smoker & Indicates the type of smoking status of the patient. Whether someone was a smoker {\textgreater}= 1 year or is currently a smoker. & \centering\arraybackslash{}- \\\hline 
nb\_cig\_packs\_year & Number of cigarettes packs/year & Number of cigarette packages consumed per year. & \centering\arraybackslash{}- \\\hline 
nb\_cigs\_day & Number of cigarettes/day & Number of cigarettes per day consumed by the patient. & \centering\arraybackslash{}- \\\hline 
lives\_with\_smokers & Lives with smoker & Whether or not the patient lives with smokers. & \centering\arraybackslash{}- \\\hline 
lived\_with\_smokers\_20\_years & Lived with smokers 20 years & Whether or not the patient has lived with smokers in the last 20 years. & \centering\arraybackslash{}- \\\hline 
\multicolumn{4}{|p{\dimexpr 1\linewidth-2\tabcolsep-2\arrayrulewidth}|}{\centering\arraybackslash{}\centering\arraybackslash{}\textbf{General Features}} \\\hline 
age & Age & Patient age at diagnosis, approximated with heuristic based on first available date for missing values. & \centering\arraybackslash{}- \\\hline 
race & Race & Race of the patient. & \centering\arraybackslash{}- \\\hline 
previous\_cancer\_type & Previous cancer type & Type of oncological antecedent experienced by the patient. & \centering\arraybackslash{}- \\\hline 
family\_lung\_cancer & Number of family members with lung cancer & Number of family relatives with lung cancer. & \centering\arraybackslash{}- \\\hline 
family\_other\_cancer & Number of family members with other cancer & Number of family relatives with other types of cancer. & \centering\arraybackslash{}- \\\hline 
gender & Gender & Gender of the patient. & \centering\arraybackslash{}- \\\hline 
\multicolumn{4}{|p{\dimexpr 1\linewidth-2\tabcolsep-2\arrayrulewidth}|}{\centering\arraybackslash{}\centering\arraybackslash{}\textbf{Diagnosis Features}} \\\hline 
histology\_grade & Histology Grade & Histological degree presented by the patient's tumor. & \centering\arraybackslash{}- \\\hline 
tumor\_histology & Tumor Histology & Type of histology associated to the patient's tumour. & \centering\arraybackslash{}- \\\hline 
diagnosis\_ecog & ECOG & Value describing the patient's performance status (in relationship with the scale indicated in the "scale" column). & \centering\arraybackslash{}- \\\hline 
diagnosis\_synchronous\_tumours & Synchronous tumors & Whether or not the patient has synchronous tumours. & \centering\arraybackslash{}- \\\hline 
adeno\_histology & Histology subtype: adeno histology & - & \centering\arraybackslash{}- \\\hline 
histology\_subtype & Histology Subtype & - & \centering\arraybackslash{}- \\\hline 
biomarker@diagnosis\_ALK\_IHQ & ALK IHQ & IHQ result of the ALK study. & \centering\arraybackslash{}- \\\hline 
biomarker@diagnosis\_PD-L1 & PD-L1 & PDL1 result. & \centering\arraybackslash{}- \\\hline 
biomarker@diagnosis\_type & Biomarker type & Marker studied. & \centering\arraybackslash{}- \\\hline 
biomarker@diagnosis\_EGFR\_negative & EGFR negative & Whether or not the result of this biomarker's study was negative. & \centering\arraybackslash{}- \\\hline 
\end{tabularx}
\end{table}

\begin{table}[H]
\begin{tabularx}{\textwidth}{|
p{\dimexpr 0.403\linewidth-2\tabcolsep-2\arrayrulewidth}|
p{\dimexpr 0.223\linewidth-2\tabcolsep-\arrayrulewidth}|
p{\dimexpr 0.243\linewidth-2\tabcolsep-\arrayrulewidth}|
p{\dimexpr 0.131\linewidth-2\tabcolsep-\arrayrulewidth}|} \hline 
\textbf{Features} & \textbf{Name} & \textbf{Description} & \centering\arraybackslash{}\textbf{Filter} \\\hline 
\multicolumn{4}{|p{\dimexpr 1\linewidth-2\tabcolsep-2\arrayrulewidth}|}{\centering\arraybackslash{}\centering\arraybackslash{}\textbf{Symptoms}} \\\hline 
symptom\_asymptomatic & Asymptomatic & Whether the patient was asymptomatic. & \centering\arraybackslash{}- \\\hline 
symptom\_cough & Cough & \multirow[t]{8}{=}{Whether the patient presented respective symptom.} & \centering\arraybackslash{}- \\\cline{1-2}\cline{4-4}
symptom\_pain & Pain &  & \centering\arraybackslash{}- \\\cline{1-2}\cline{4-4}
symptom\_dyspnea & Dyspnea &  & \centering\arraybackslash{}- \\\cline{1-2}\cline{4-4}
symptom\_hemoptits & Hemoptits &  & \centering\arraybackslash{}- \\\cline{1-2}\cline{4-4} 
symptom\_weight\_loss & Weight Loss &  & \centering\arraybackslash{}- \\\cline{1-2}\cline{4-4} 
symptom\_anorexia & Anorexia &  & \centering\arraybackslash{}- \\\cline{1-2}\cline{4-4}
symptom\_astenia & Astenia &  & \centering\arraybackslash{}- \\\cline{1-2}\cline{4-4}
symptom\_others & Others &  & \centering\arraybackslash{}- \\\cline{1-2}\cline{4-4} 
\multicolumn{4}{|p{\dimexpr 1\linewidth-2\tabcolsep-2\arrayrulewidth}|}{\centering\arraybackslash{}\centering\arraybackslash{}\textbf{Radiotherapy Features}} \\\hline 
radiotherapy@t1\_type & Type & Type of radiotherapy conducted. & \centering\arraybackslash{}- \\\hline 
radiotherapy@t1\_area & Area & Area radiated during the readiotherapy treatment. & \centering\arraybackslash{}- \\\hline 
radiotherapy@t1\_dose & Dose & Dosage of radiotherapy applied to the patient (measured in Gray units). & \centering\arraybackslash{}- \\\hline 
radiotherapy@t1\_fractioning & Fractioning & Number describing the amount of parts in which the radiotherapy dose was divided. & \centering\arraybackslash{}- \\\hline 
radiotherapy@t1\_duration\_days & Duration {[}days{]} & Days elapsed between start of radiotherapy and end of radiotherapy. & \centering\arraybackslash{}- \\\hline 
\multicolumn{4}{|p{\dimexpr 1\linewidth-2\tabcolsep-2\arrayrulewidth}|}{\centering\arraybackslash{}\centering\arraybackslash{}\textbf{Chemotherapy Features}} \\\hline 
chemotherapy@t1\_start\_time & Start Time {[}months{]} & Chemotherapy starts in months since the first recorded event in patient history. & \centering\arraybackslash{}- \\\hline 
chemotherapy@t1\_regimen & Regimen & Class of chemotherapy process conducted on the patient. & \centering\arraybackslash{}- \\\hline 
\multicolumn{4}{|p{\dimexpr 1\linewidth-2\tabcolsep-2\arrayrulewidth}|}{\centering\arraybackslash{}\centering\arraybackslash{}\textbf{Surgery Features}} \\\hline 
surgery@t1\_procedure1 & Procedure 1 & 1st surgery iteration in which the surgery was conducted. & \centering\arraybackslash{}- \\\hline 
surgery@t1\_time & Time & Surgery time in months since the first recorded event in patient history. & \centering\arraybackslash{}- \\\hline 
surgery@t1\_type & Type & Type of surgery perfomed on the patient. & \centering\arraybackslash{}- \\\hline 
surgery@t1\_resection\_grade & Resection Grade & Degree of resection associated to the surgery, & \centering\arraybackslash{}- \\\hline 
surgery@t1\_t\_stage & Stage T & T stage of the patient. Measured before the conduction of surgery. & \centering\arraybackslash{}- \\\hline 
surgery@t1\_n\_stage & Stage N & N stage of the patient. Measured before the conduction of surgery. & \centering\arraybackslash{}- \\\hline 
surgery@t1\_perioperative\_death & Perioperative Death & Whether or not perioperative death took place within 30 days after surgery. & \centering\arraybackslash{}- \\\hline 
\end{tabularx}
\end{table}

\begin{table}
\begin{tabularx}{\textwidth}{|
p{\dimexpr 0.343\linewidth-2\tabcolsep-2\arrayrulewidth}|
p{\dimexpr 0.23\linewidth-2\tabcolsep-\arrayrulewidth}|
p{\dimexpr 0.427\linewidth-2\tabcolsep-\arrayrulewidth}|} \hline 
\multicolumn{3}{|p{\dimexpr 1\linewidth-2\tabcolsep-2\arrayrulewidth}|}{\centering\arraybackslash{}\centering\arraybackslash{}\textbf{Features with all values missing for the selected cohort (NOT included in the model)}} \\\hline 
mesothelioma & Histology subtype: mesothelioma & \multirow[t]{2}{=}{\centering{}-}  \\\cline{1-2}
small\_cell\_carcinom & Histology subtype: small cell carcinom &  \\\hline 
\end{tabularx}
\caption*{\textit{Table 5 Comprehensive list of features.}}
\end{table}

\section*{E Class-Based Evaluation }

\textit{Tabular Models (averaged on all 10 folds):}
\begin{table}[H]
\begin{tabularx}{\textwidth}{|
p{\dimexpr 0.502\linewidth-2\tabcolsep-2\arrayrulewidth}|
p{\dimexpr 0.498\linewidth-2\tabcolsep-\arrayrulewidth}|} \hline 
\raggedright\arraybackslash{}\textbf{Baseline} & \raggedright\arraybackslash{}\textbf{Random Forest} \\\hline 
\raggedright\arraybackslash{}Accuracy: 0.514${\pm}$0.041 & \raggedright\arraybackslash{}Accuracy: 0.7628${\pm}$0.031 \\\hline 
\raggedright\arraybackslash{}No-Relapse: & \raggedright\arraybackslash{}No-Relapse: \\\hline 
\raggedright\arraybackslash{}precision: 0.6384${\pm}$0.038 & \raggedright\arraybackslash{}precision: 0.7938${\pm}$0.026 \\\hline 
\raggedright\arraybackslash{}recall: 0.5207${\pm}$0.061 & \raggedright\arraybackslash{}recall: 0.8414${\pm}$0.042 \\\hline 
\raggedright\arraybackslash{}f1-score: 0.572${\pm}$0.045 & \raggedright\arraybackslash{}f1-score: 0.8162${\pm}$0.026 \\\hline 
\raggedright\arraybackslash{}support: 87.0${\pm}$0.0 & \raggedright\arraybackslash{}support: 87.0${\pm}$0.0 \\\hline 
\raggedright\arraybackslash{}Relapse: & \raggedright\arraybackslash{}Relapse: \\\hline 
\raggedright\arraybackslash{}precision: 0.3847${\pm}$0.042 & \raggedright\arraybackslash{}precision: 0.7055${\pm}$0.051 \\\hline 
\raggedright\arraybackslash{}recall: 0.5032${\pm}$0.072 & \raggedright\arraybackslash{}recall: 0.6302${\pm}$0.063 \\\hline 
\raggedright\arraybackslash{}f1-score: 0.4347${\pm}$0.047 & \raggedright\arraybackslash{}f1-score: 0.6636${\pm}$0.048 \\\hline 
\raggedright\arraybackslash{}support: 51.7${\pm}$0.46 & \raggedright\arraybackslash{}support: 51.7${\pm}$0.46 \\\hline 
\raggedright\arraybackslash{}Macro Average: & \raggedright\arraybackslash{}Macro Average: \\\hline 
\raggedright\arraybackslash{}precision: 0.5116${\pm}$0.04 & \raggedright\arraybackslash{}precision: 0.7496${\pm}$0.034 \\\hline 
\raggedright\arraybackslash{}recall: 0.5119${\pm}$0.042 & \raggedright\arraybackslash{}recall: 0.7358${\pm}$0.034 \\\hline 
\raggedright\arraybackslash{}f1-score: 0.5034${\pm}$0.04 & \raggedright\arraybackslash{}f1-score: 0.7399${\pm}$0.035 \\\hline 
\raggedright\arraybackslash{}support: 138.7${\pm}$0.46 & \raggedright\arraybackslash{}support: 138.7${\pm}$0.46 \\\hline 
\raggedright\arraybackslash{}Weighted Average: & \raggedright\arraybackslash{}Weighted Average: \\\hline 
\raggedright\arraybackslash{}precision: 0.5439${\pm}$0.039 & \raggedright\arraybackslash{}precision: 0.7608${\pm}$0.031 \\\hline 
\raggedright\arraybackslash{}recall: 0.514${\pm}$0.041 & \raggedright\arraybackslash{}recall: 0.7628${\pm}$0.031 \\\hline 
\raggedright\arraybackslash{}f1-score: 0.5208${\pm}$0.04 & \raggedright\arraybackslash{}f1-score: 0.7594${\pm}$0.032 \\\hline 
\raggedright\arraybackslash{}support: 138.7${\pm}$0.46 & \raggedright\arraybackslash{}support: 138.7${\pm}$0.46 \\\hline 
\raggedright\arraybackslash{} & \raggedright\arraybackslash{} \\\hline 
\raggedright\arraybackslash{}\textbf{Gradient Boosting} & \raggedright\arraybackslash{}\textbf{Logistic Regression} \\\hline 
\raggedright\arraybackslash{}Accuracy: 0.7505${\pm}$0.031 & \raggedright\arraybackslash{}Accuracy: 0.7181${\pm}$0.038 \\\hline 
\raggedright\arraybackslash{}No-Relapse: & \raggedright\arraybackslash{}No-Relapse: \\\hline 
\raggedright\arraybackslash{}precision: 0.8041${\pm}$0.03 & \raggedright\arraybackslash{}precision: 0.7984${\pm}$0.035 \\\hline 
\raggedright\arraybackslash{}recall: 0.7989${\pm}$0.055 & \raggedright\arraybackslash{}recall: 0.7379${\pm}$0.046 \\\hline 
\raggedright\arraybackslash{}f1-score: 0.8${\pm}$0.03 & \raggedright\arraybackslash{}f1-score: 0.7662${\pm}$0.033 \\\hline 
\raggedright\arraybackslash{}support: 87.0${\pm}$0.0 & \raggedright\arraybackslash{}support: 87.0${\pm}$0.0 \\\hline 
\raggedright\arraybackslash{}Relapse: & \raggedright\arraybackslash{}Relapse: \\\hline 
\raggedright\arraybackslash{}precision: 0.6685${\pm}$0.047 & \raggedright\arraybackslash{}precision: 0.6097${\pm}$0.047 \\\hline 
\raggedright\arraybackslash{}recall: 0.6687${\pm}$0.076 & \raggedright\arraybackslash{}recall: 0.6843${\pm}$0.067 \\\hline 
\raggedright\arraybackslash{}f1-score: 0.6651${\pm}$0.045 & \raggedright\arraybackslash{}f1-score: 0.6435${\pm}$0.049 \\\hline 
\raggedright\arraybackslash{}support: 51.7${\pm}$0.46 & \raggedright\arraybackslash{}support: 51.7${\pm}$0.46 \\\hline 
\raggedright\arraybackslash{}Macro Average: & \raggedright\arraybackslash{}Macro Average: \\\hline 
\raggedright\arraybackslash{}precision: 0.7363${\pm}$0.031 & \raggedright\arraybackslash{}precision: 0.704${\pm}$0.038 \\\hline 
\raggedright\arraybackslash{}recall: 0.7338${\pm}$0.034 & \raggedright\arraybackslash{}recall: 0.7111${\pm}$0.04 \\\hline 
\raggedright\arraybackslash{}f1-score: 0.7326${\pm}$0.033 & \raggedright\arraybackslash{}f1-score: 0.7048${\pm}$0.039 \\\hline 
\raggedright\arraybackslash{}support: 138.7${\pm}$0.46 & \raggedright\arraybackslash{}support: 138.7${\pm}$0.46 \\\hline 
\raggedright\arraybackslash{}Weighted Average: & \raggedright\arraybackslash{}Weighted Average: \\\hline 
\raggedright\arraybackslash{}precision: 0.7535${\pm}$0.029 & \raggedright\arraybackslash{}precision: 0.728${\pm}$0.037 \\\hline 
\raggedright\arraybackslash{}recall: 0.7505${\pm}$0.031 & \raggedright\arraybackslash{}recall: 0.7181${\pm}$0.038 \\\hline 
\raggedright\arraybackslash{}f1-score: 0.7498${\pm}$0.031 & \raggedright\arraybackslash{}f1-score: 0.7205${\pm}$0.037 \\\hline 
\raggedright\arraybackslash{}support: 138.7${\pm}$0.46 & \raggedright\arraybackslash{}support: 138.7${\pm}$0.46 \\\hline 
\end{tabularx}
\end{table}

\begin{table}[H]
\begin{tabularx}{\textwidth}{|
p{\dimexpr 0.502\linewidth-2\tabcolsep-2\arrayrulewidth}|p{\dimexpr 0.498\linewidth-2\tabcolsep-\arrayrulewidth}|} 
\hline 
\raggedright\arraybackslash{}\textbf{Graph ML: ComplEx-N3} &   \\\hline 
\raggedright\arraybackslash{}Accuracy: 0.685 & \\\hline 
\raggedright\arraybackslash{}No-Relapse: &\\\hline 
\raggedright\arraybackslash{}precision: 0.6609 &\\\hline 
\raggedright\arraybackslash{}recall: 0.76  &\\\hline 
\raggedright\arraybackslash{}f1-score: 0.707  &\\\hline 
\raggedright\arraybackslash{}support: 100 &\\\hline 
\raggedright\arraybackslash{}Relapse:  &\\\hline 
\raggedright\arraybackslash{}precision: 0.7176  &\\\hline 
\raggedright\arraybackslash{}recall: 0.61  &\\\hline 
\raggedright\arraybackslash{}f1-score: 0.6595  &\\\hline 
\raggedright\arraybackslash{}support: 100  &\\\hline 
\raggedright\arraybackslash{}Macro Average: precision: 0.6893  &\\\hline 
\raggedright\arraybackslash{}recall: 0.685 & \\\hline 
\raggedright\arraybackslash{}f1-score: 0.6832 & \\\hline 
\raggedright\arraybackslash{}support: 200 & \\\hline 
\raggedright\arraybackslash{}Weighted Average: & \\\hline 
\raggedright\arraybackslash{}precision: 0.6893 & \\\hline 
\raggedright\arraybackslash{}recall: 0.685 & \\\hline 
\raggedright\arraybackslash{}f1-score: 0.6832 & \\\hline 
\raggedright\arraybackslash{}support: 200 & \\\hline 
\end{tabularx}
\caption{\textit{Table 6 Class-based predictions for models.}}
\end{table}

\section*{F Confusion Matrices for Models}

\begin{figure}[H]
\includegraphics[width=0.7\textwidth]{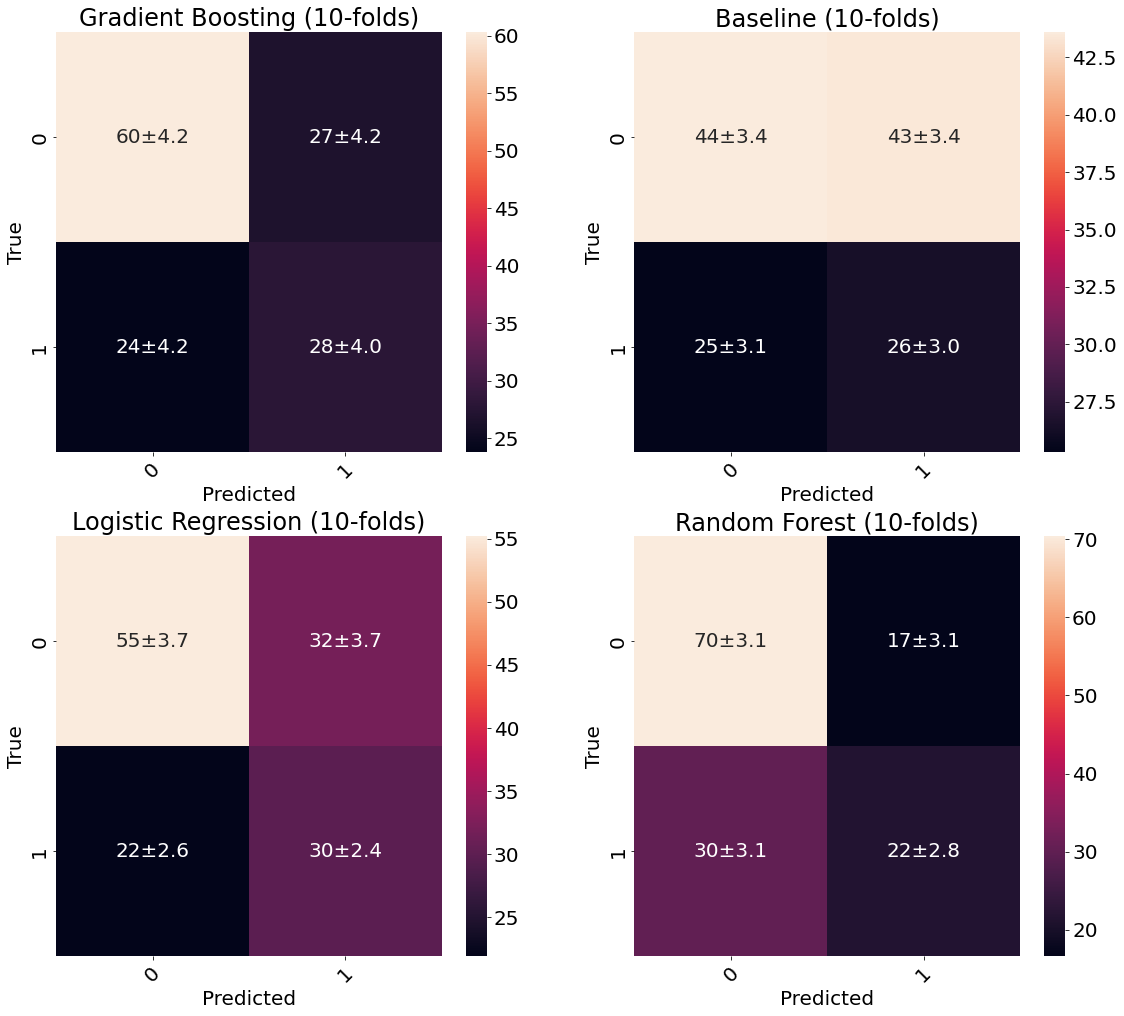}\caption{\textit{Figure 19 Diagnosis features}}

\label{fig:11}
\end{figure}

\begin{figure}
\includegraphics[width=0.9\textwidth]{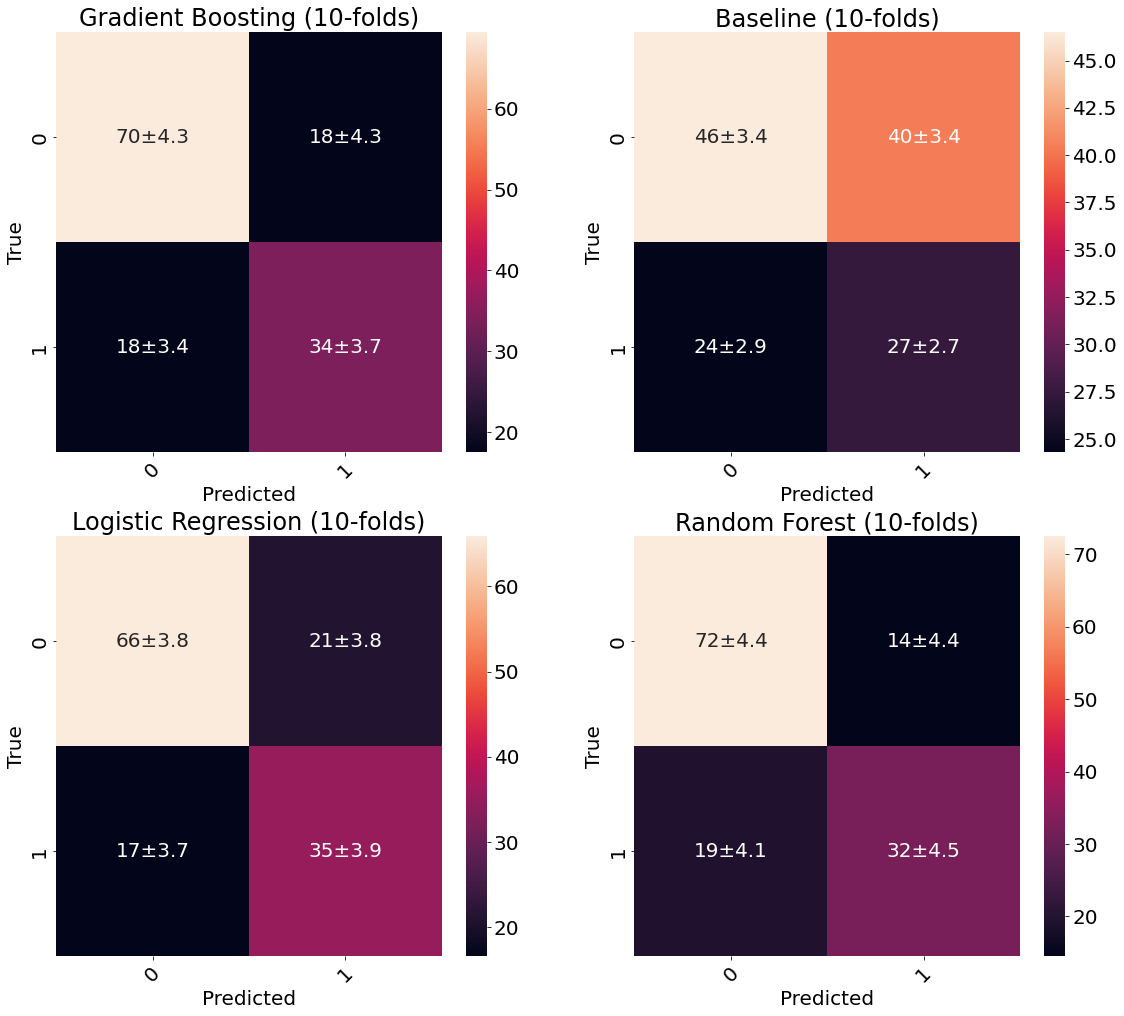}\caption{\textit{Figure 16 All features}}
\label{fig:8}
\end{figure}

\begin{figure}[H]
\textit{\includegraphics[width=0.45\textwidth]{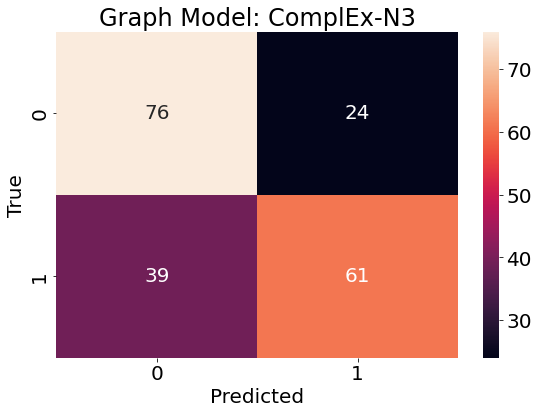}}   
\caption{\textit{Figure 20 Graph ML ComplEx-N3 confusion matrix.}}
\label{fig:12}
\end{figure}

\begin{figure}
\textit{\includegraphics[width=0.67\textwidth]{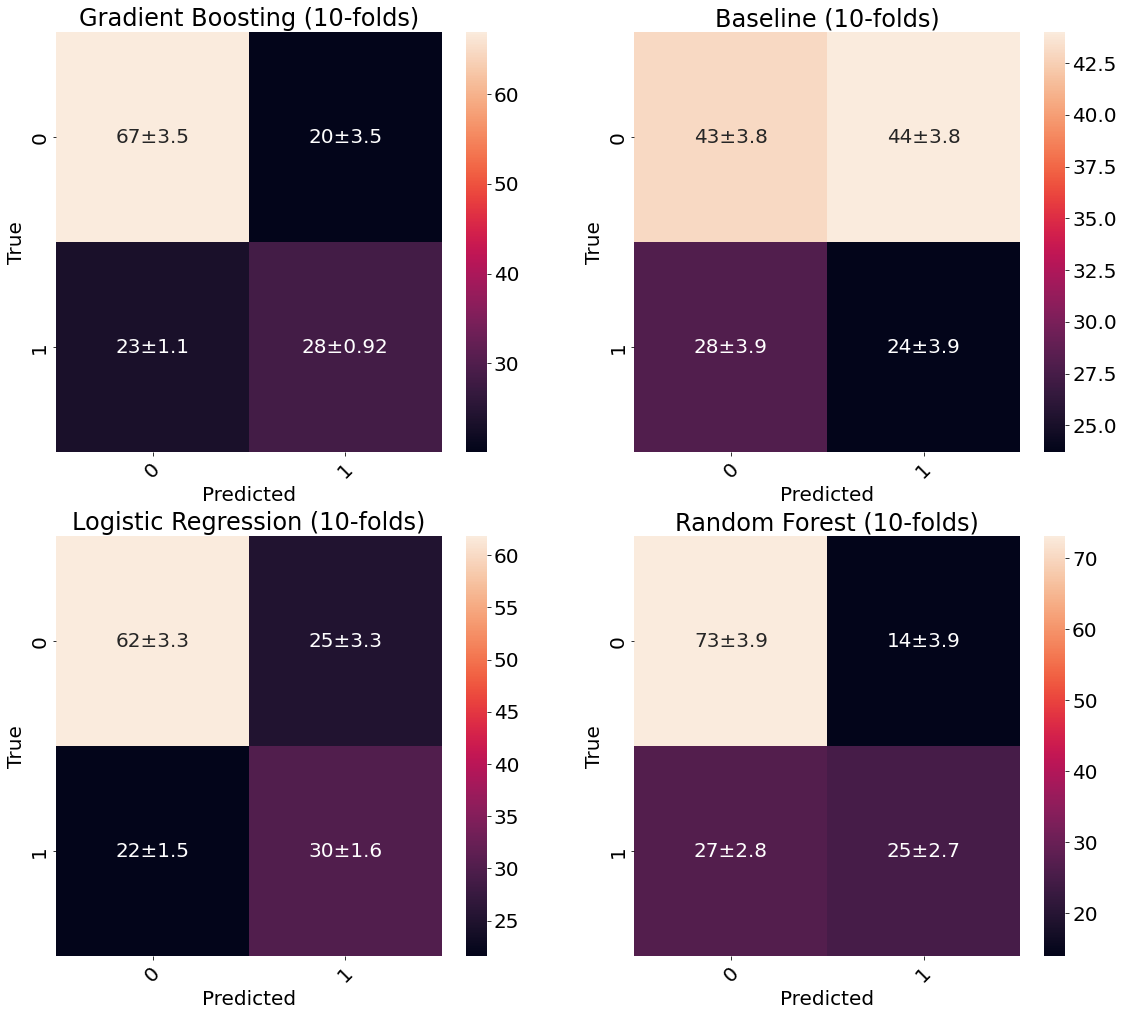}}\caption{\textit{Figure 18 Up to surgery features}}
\label{fig:10}
\end{figure}

\begin{figure}
\includegraphics[width=0.67\textwidth]{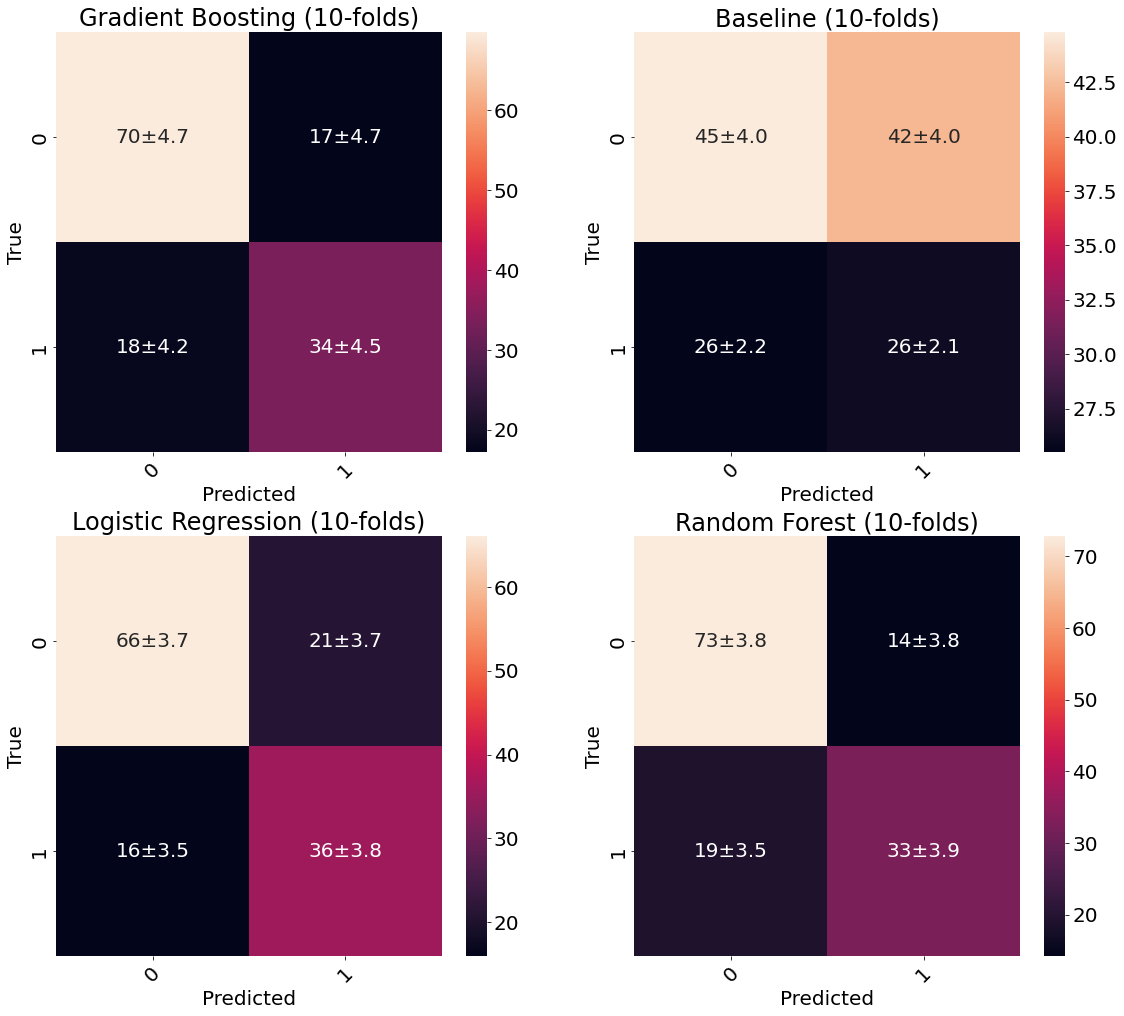}\caption{\textit{Figure 17 Up to chemotherapy features}}
\label{fig:9}
\end{figure}

\section*{G Receiver Operating Characteristic (ROC) Curves for Models}

\begin{figure}
\includegraphics[width=0.85\textwidth]{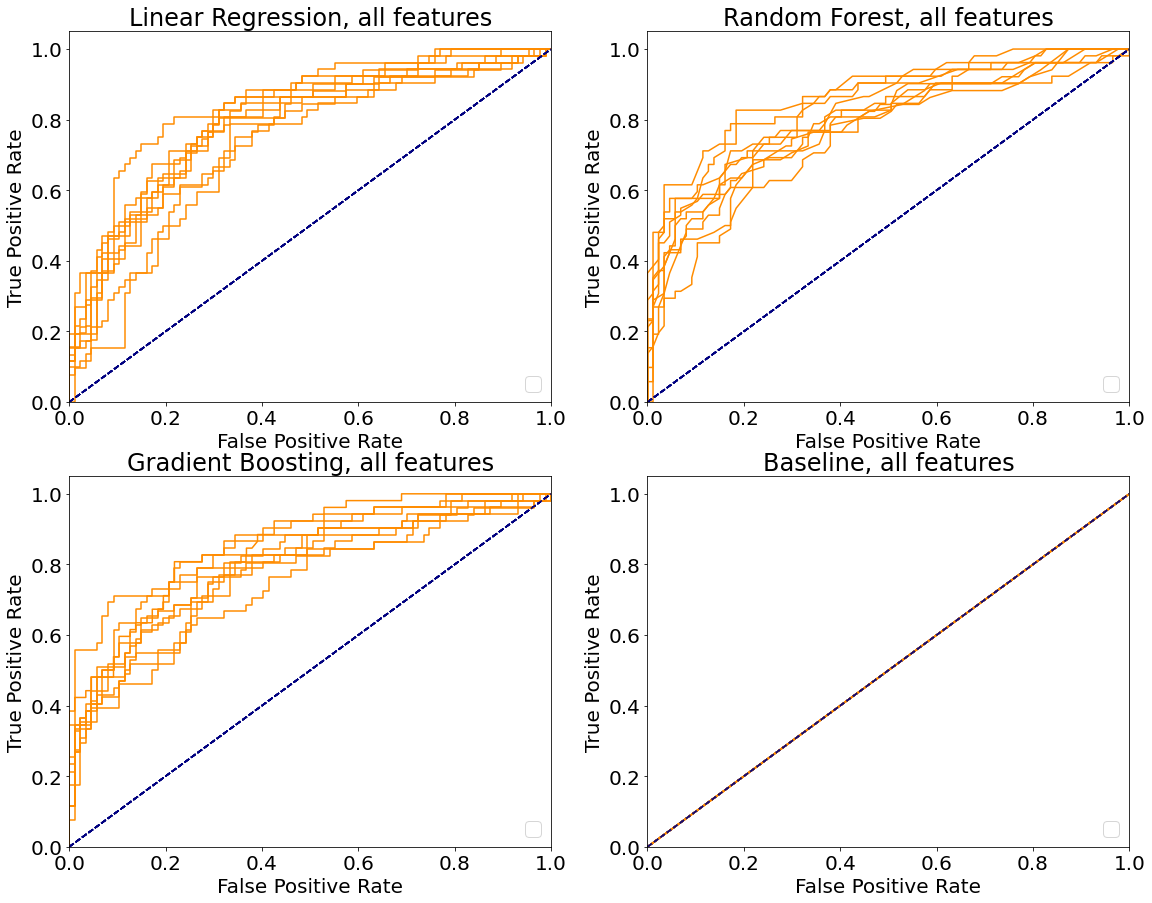}
\caption{\textit{Figure 21 ROC Curves, all features.}}
\label{fig:13}
\end{figure}

\begin{figure}
\includegraphics[width=0.65\textwidth]{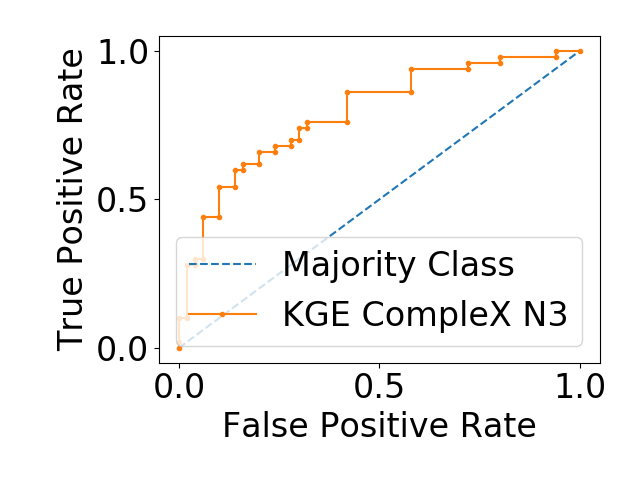}\caption{\textit{Figure 25 ROC Curve,} \textit{Graph ML model ComplEx-N3, all features.}}
\label{fig:17}
\end{figure}

\begin{figure}
\includegraphics[width=0.75\textwidth]{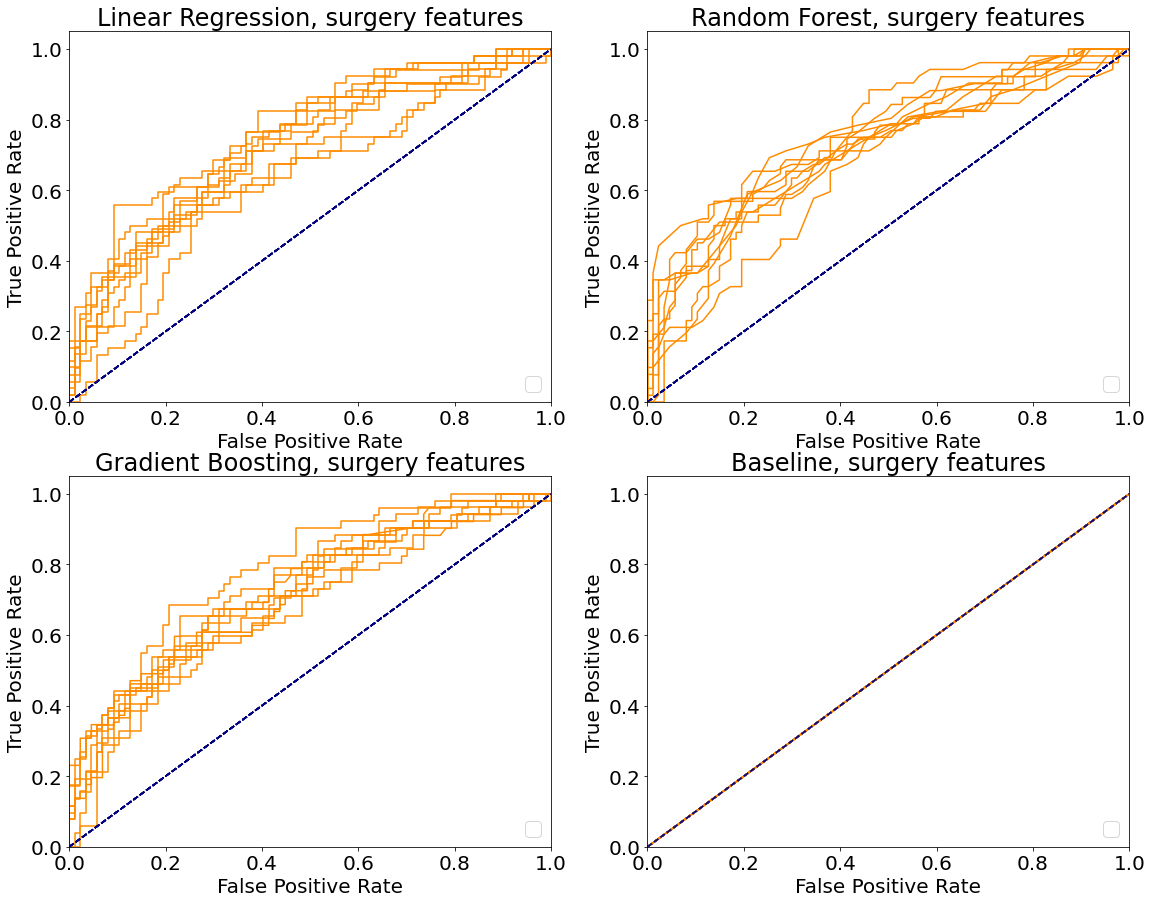}
\caption{\textit{Figure 22 ROC Curves, surgery features.}}
\label{fig:14}
\end{figure}

\begin{figure}
\includegraphics[width=0.75\textwidth]{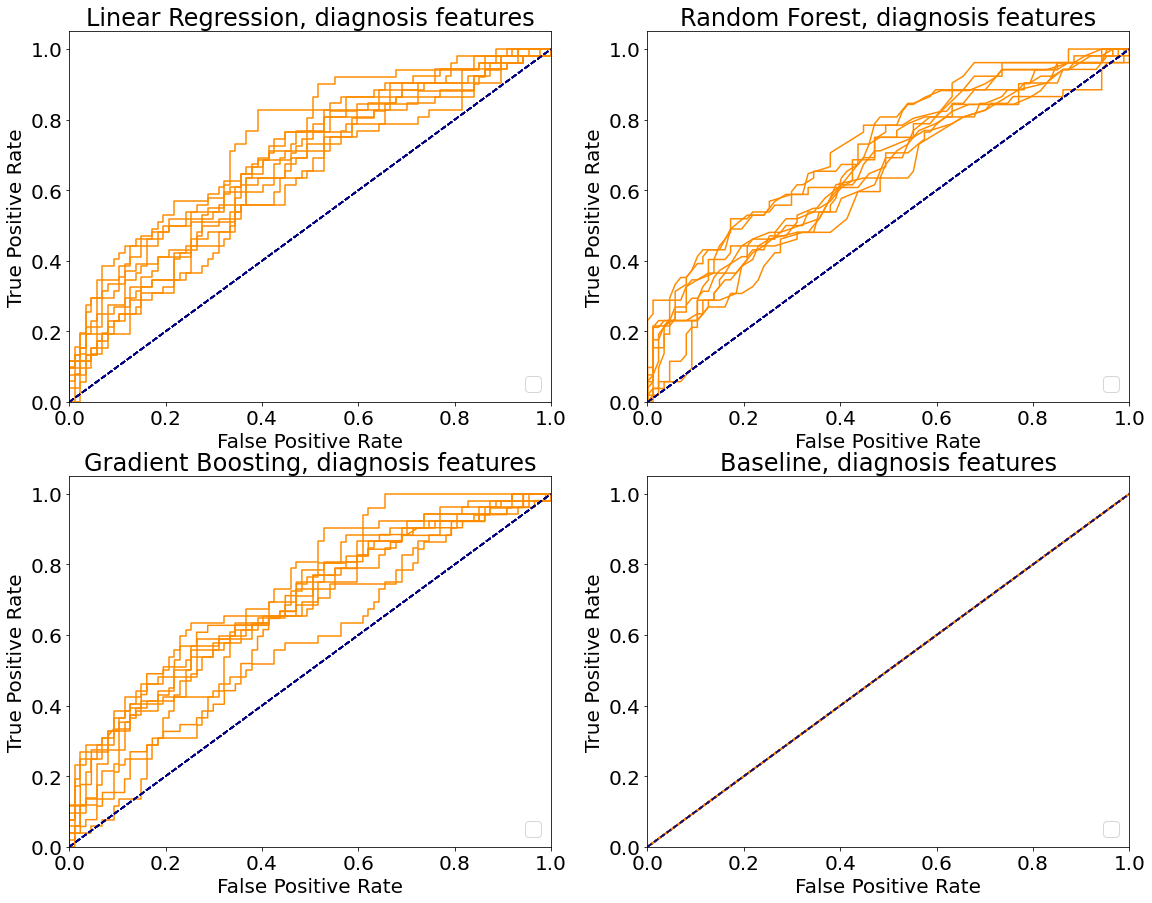}\caption{\textit{Figure 23 ROC Curves, diagnosis features.}}
\label{fig:15}
\end{figure}

\begin{figure}
\includegraphics[width=0.85\textwidth]{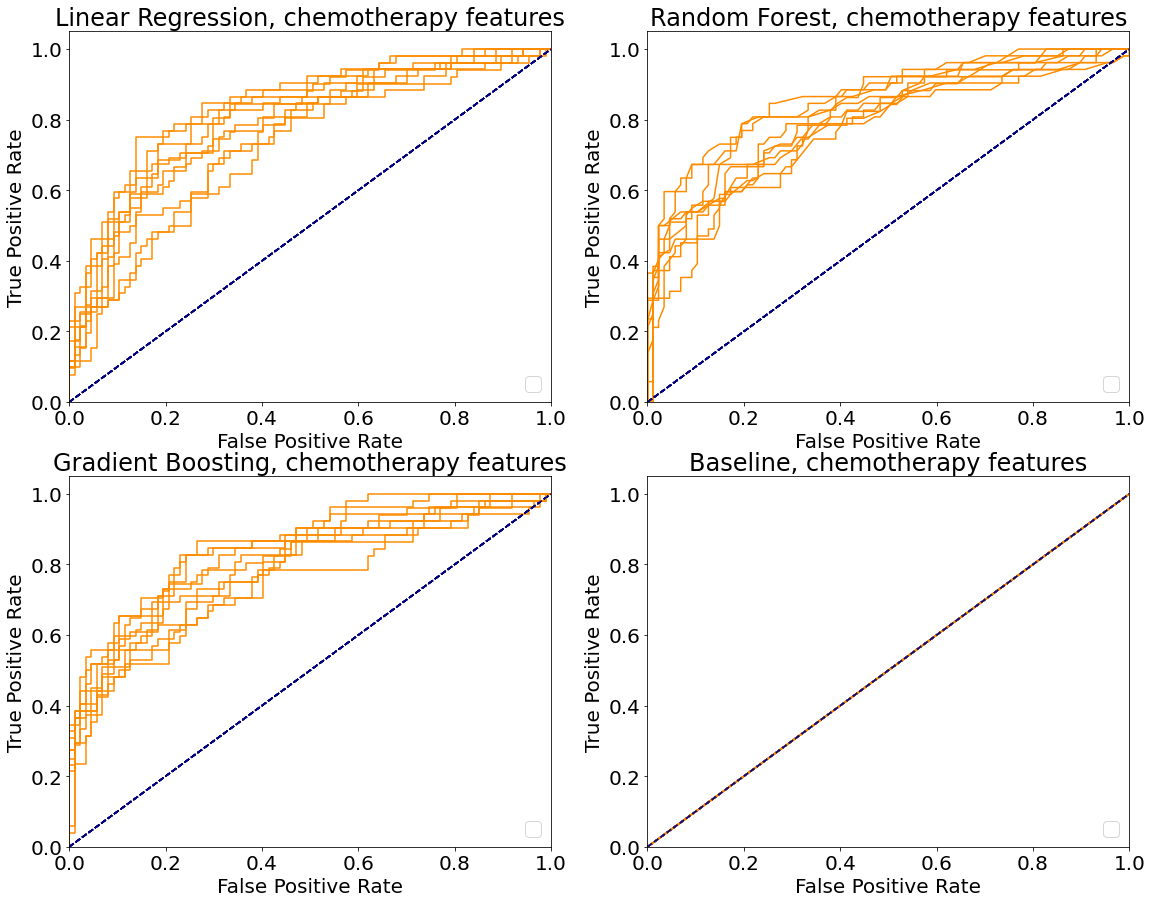}\caption{\textit{Figure 24 ROC Curves, chemotherapy features.}}
\label{fig:16}
\end{figure}

\section*{H Waterfall Plots for Tabular Models}

\begin{figure}
\includegraphics[width=1\textwidth]{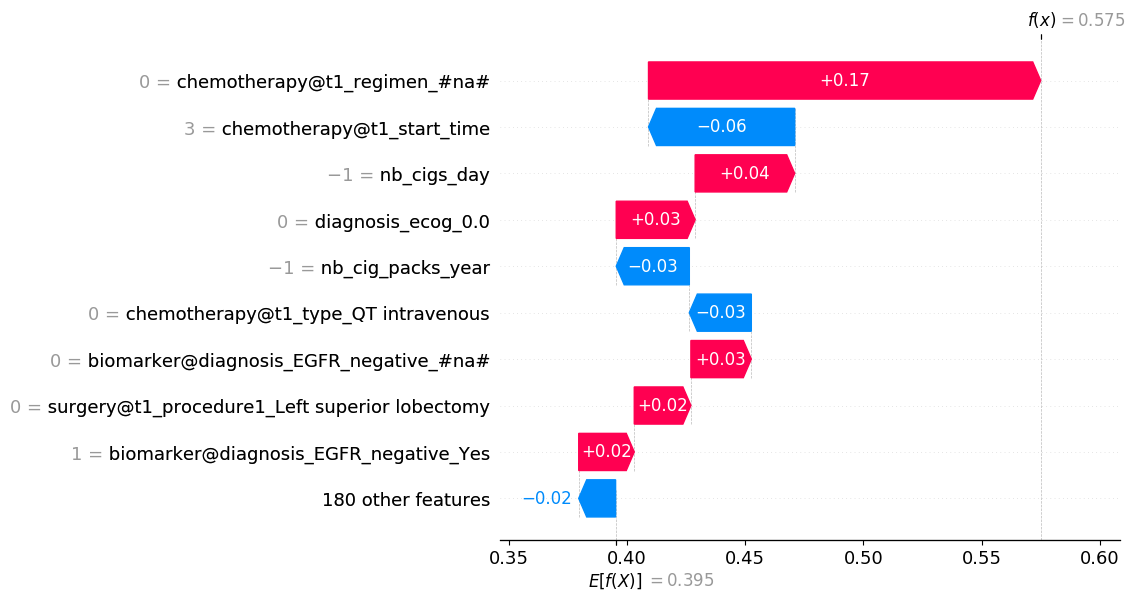}
\caption{\textit{Figure 26 Gradient Boosting, up-to treatment}}
\label{fig:18}
\end{figure}

\begin{figure}
\includegraphics[width=1\textwidth]{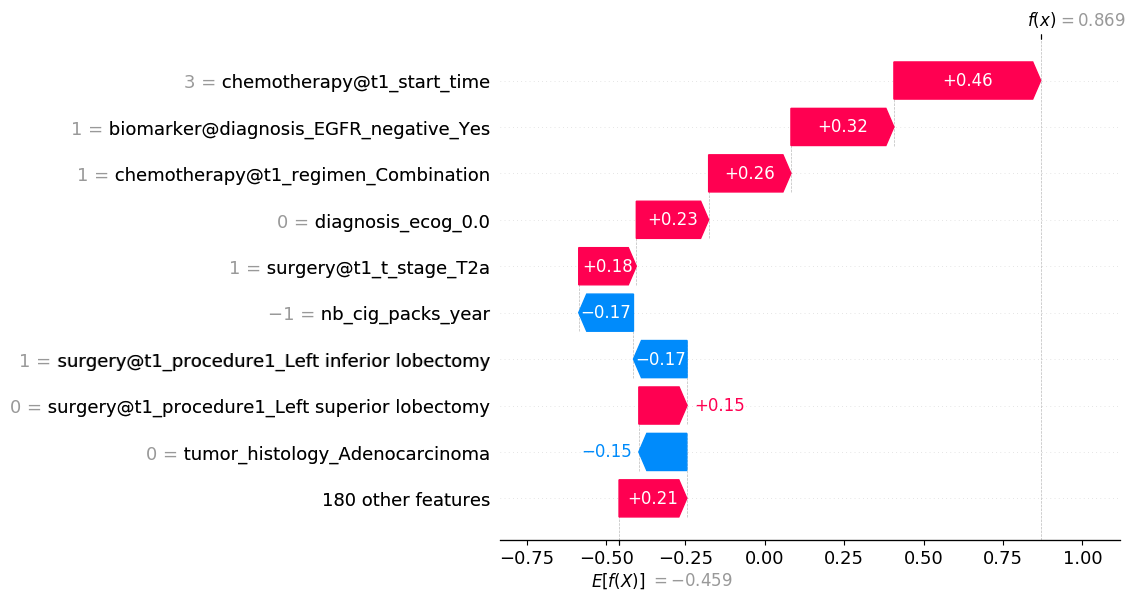}\caption{\textit{Figure 27 Logistic Regression, up-to treatment}}
\label{fig:19}
\end{figure}

\begin{figure}
\includegraphics[width=1\textwidth]{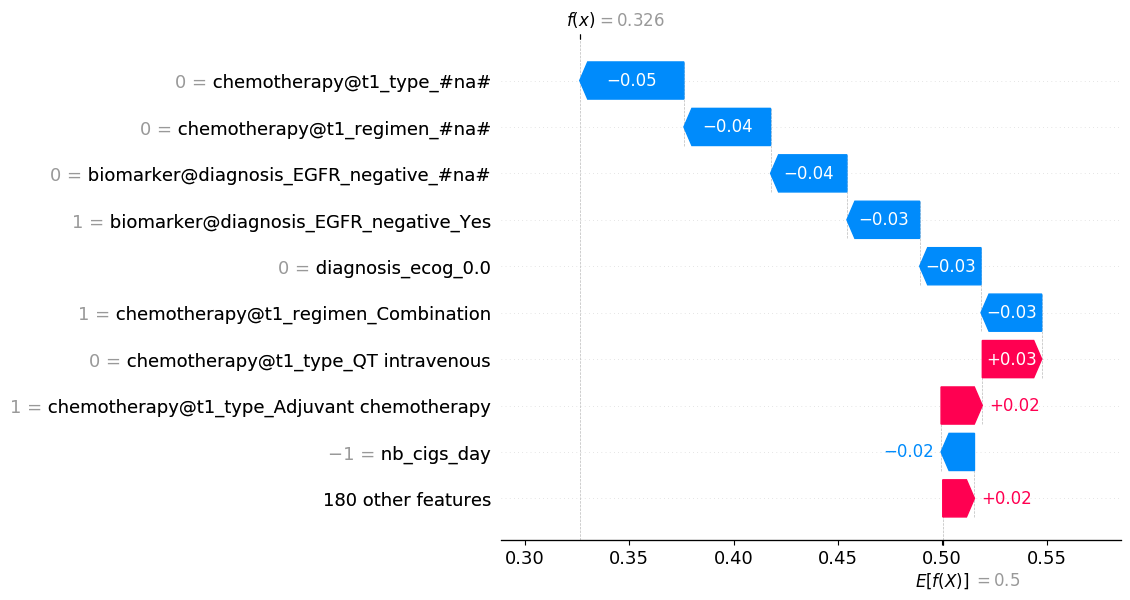}\caption{\textit{Figure 28 Random Forest, up-to treatment}}
\label{fig:20}
\end{figure}

\section*{I Metrics}

\textbf{Binary Evaluation Metrics}

$TP$ - True Positives

$FP$ - False Positives

$TN$ - True Negatives

$FN$ - False Negatives

$TPR=\frac{TP}{TP+FN}$

$\textit{Accuracy}=\frac{TP+TN}{TP+TN+FP+FN}$ 

$P/\textit{Precision}=\frac{TP}{TP+FP}$ 

$R/\textit{Recall}=\frac{TP}{TP+FN}$

$F1=\frac{2PR}{P+R}$

AUC-ROC/ Area under the Receiver Operating Characteristics curve.

AUC-ROC $=\int TPR d(FPR)$ 

Average Precision/ AP/ AUC-PR/ Area under Precision Recall curve.

AUC-PR $=\int P d(R)$

\textbf{Knowledge Graph Embedding Evaluation Metrics}

$Hits@N = \sum_{i=0}^{|Q|}1$ if $rank_{(s,p,o)_i} <= N$

MRR/Mean Reciprocal Rank MRR $= \frac{1}{|Q|} \sum_{i=1}^{|Q|} \frac{1}{rank_{(s,p,o)_i}}$

\section*{J Model Parameters}

Best Hyperparameters for Tabular Models:
\begin{itemize}
\item Random Forest: n estimators = 170
\item Gradient Boosting: n estimators = 150
\item Logistic Regression: solver=liblinear, C=0.0774, penalty=l2
\end{itemize}
\section*{H Time-based Relapse Prediction*}

We have a \textbf{general relapse timestamp} available in our dataset. So, we selected the cut-off time for relapse prediction. We made the interval of 3,6,9 and 12 months as a cut-off time and censored the relapse event based on these time stamps. For example, if we took a cut-off time of \textbf{6 months} and a general relapse timestamp of this patient is \textbf{8 months} then we relabel this patient as non-relapse. This way for every cut-off time the distribution of patient relapse events changes. 

We trained three different state-of-the-art survival models namely \textit{Cox Proportional Hazard,} \textit{Random Forest Survival Model} and \textit{Gradient Boosted Model} and performed \textbf{10-fold cross validation} and evaluated the model based on the Concordance Index (CI). CI provides the rank correlation between predicted risk scores and observed time point. Note that this metric cannot be compared across the suite of machine learning models reported in the article due to insufficient reliable time-stamp data, therefore we believe it is meaningful to only elaborate on it in a supplement, as a stepping stone for possible future work. 

The evaluation of the survival model is provided below:
\begin{figure}
\includegraphics[width=1\textwidth]{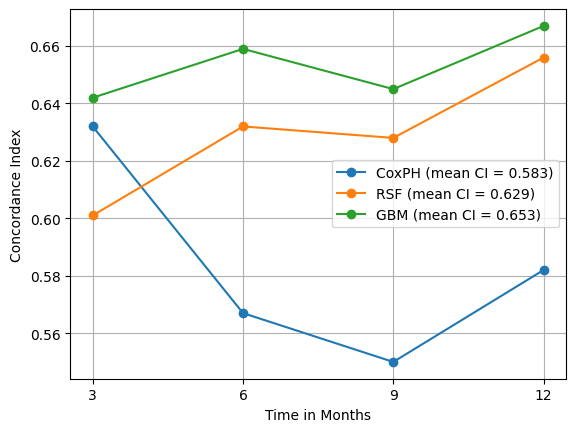}\caption{Figure 1: Comparison of the Survival Model for Time-based Relapse prediction}
\label{fig:21}
\end{figure}
From the figure, we can see that the GBM performed better in comparison to CoXPH and RSF during the 3 to 12 months interval for relapse prediction. Among three survival models CoXPH performed worse. It is due to the fact that CoXPH is the linear survival modelling technique and operates on multiple assumptions {[}1{]}. However, GBM and RSF capture the non-linear relationship between the predictors and have higher predictive performance in these datasets. On average, GBM outperformed RSF by 3.81\% and CoxPH by 12.01\%. 

In the above experiment, we have used diagnostic features in the survival model. It includes symptoms, gender, age, sex, previous cancer, tumor stages and comorbodity. We have not included the time based features like chemotherapy, radiotherapy, biomarker and surgery information in the survival models, since we lack sufficient number of reliable time-stamps for these features in the collected patient data. Therefore the primary focus of our work was developing supervised machine learning and knowledge graph embedding models without time constraint. That let us utilise all available features and make more accurate predictions that are already clinically useful as they let the clinicians classify the patients into high and low risk cohorts based on the predicted relapse likelihood scores, which was the clinical end point of this study from the very beginning. Follow-up models and their evaluation for time based relapse prediction with time based features or predictors will, however, be addressed in our future work as a separate publication once we manage to collect further data.

*Courtesy of Mohan Timilsina

References:

{[}1{]} Hess, Kenneth R. "Graphical methods for assessing violations of the proportional hazards assumption in Cox regression." \textit{Statistics in medicine} 14.15 (1995): 1707-1723.